\documentclass[10pt,twocolumn,letterpaper]{article}

\usepackage[pagenumbers]{arxiv} %

\usepackage[most]{tcolorbox}
\definecolor{cargo}{RGB}{254,148,3}
\definecolor{frozen}{RGB}{61,156,188}
\definecolor{bedc}{RGB}{255,187,120}
\definecolor{wallc}{RGB}{174,199,232}
\newcommand{\chip}[1]{\textcolor{#1}{\rule[0.05ex]{0.7em}{0.7em}}}
\newcommand{\mymethod}{\textsc{CARGO}}

\newcommand{\markx}[1]{\tikz[baseline=-0.6ex]{\draw[line width=1.3pt, #1] (0,-0.09)--(0.18,0.09) (0,0.09)--(0.18,-0.09);}}
\newcommand{\markc}[1]{\tikz[baseline=-0.6ex]{\draw[line width=1.3pt, #1] (0,0)--(0.07,-0.09)--(0.22,0.11);}}
\DeclareRobustCommand{\xmark}{\markx{red}}
\DeclareRobustCommand{\cmark}{\markc{color={rgb,255:red,90;green,170;blue,60}}}

\definecolor{worldc}{RGB}{44,160,44}
\definecolor{camc}{RGB}{214,39,40}
\newcommand{\gaplab}[2]{\begin{tabular}[t]{@{}c@{}}#1\\[1pt] #2\end{tabular}}
\newcommand{\miou}[2]{\tcbox[on line, boxsep=0pt, left=2.5pt, right=2.5pt,
  top=1pt, bottom=1pt, arc=2pt, boxrule=0.4pt,
  colback=#1!10, colframe=#1!55]{#2}}
\newlength{\gw}

\newcommand{\linearprobing}{\textit{Sonata (linear)}}
\newcommand{\late}{\textit{Sonata (late blocks + linear)}}
\newcommand{\full}{\textit{Sonata (full + linear)}}

\usetikzlibrary{arrows.meta}
\tikzset{ax/.style={line width=0.6pt, -{Stealth[length=2.2pt]}}}

\newcommand{\scannetegoval}{{ScanNet-Partial (val)}}

\newcommand{\ours}{\rowcolor{cargo!10}}                       %
\newcommand{\base}[1]{\textcolor{black!50}{#1}}               %
\newcommand{\venue}[1]{{\scriptsize\textcolor{black!50}{\textsc{(#1)}}}}
\newcommand{\sbest}[1]{\underline{#1}}                        %
\DeclareRobustCommand{\lab}{\markc{black!55}}                %
\DeclareRobustCommand{\nolab}{\markx{cargo!90!black}}        %

\usepackage[skip=2pt]{caption}

\usepackage{stfloats}

\usepackage{pgfplots}
\pgfplotsset{compat=1.17}
\definecolor{arxivblue}{rgb}{0.21,0.49,0.74}
\usepackage[pagebackref,breaklinks,colorlinks,allcolors=arxivblue]{hyperref}

\usepackage{booktabs}
\usepackage{graphicx}
\usepackage[table]{xcolor}
\usepackage{multirow}

\usepackage{tikz}
\usetikzlibrary{tikzmark}
\usetikzlibrary{arrows.meta}

\usepackage[normalem]{ulem}

\usepackage{threeparttable}
\usepackage{siunitx}

\usepackage{enumitem}

\usepackage{amsmath}

\usepackage{soul}

\def\confName{arxiv}
\def\confYear{2027}

\title{\vspace{-40pt}CARGO: Camera-frame Adaptation by Relational Global distillatiOn}

\author{
  \begin{tabular}{ccc}
    {\rm Akira-Miranda Adeyomi Adeniran-Lowe}$^{1,2}$ &
    {\rm Binod Singh}$^{1,2}$ &
    {\rm Lars Arnold Dethlefsen}$^{1}$
  \end{tabular}
  \\[0.8em]
  \makebox[\textwidth][c]{%
  \begin{tabular}{cc}
    {\rm Lazaros Nalpantidis}$^{1,2}$ &
    {\rm Theodora Kontogianni}$^{1,2}$
  \end{tabular}}
  \\[0.9em]
  \makebox[\textwidth][c]{%
  \begin{tabular}{cc}
    {$^{1}$\textit{Technical University of Denmark}} &
    {$^{2}$\textit{Pioneer Center for Artificial Intelligence}}
  \end{tabular}}
  \\
  \makebox[\textwidth][c]{%
    {\href{https://aka-lowe.github.io/cargo/}{cargo.github.io}}
  }
}

\begin{document}
\twocolumn[{%
\renewcommand\twocolumn[1][]{#1}%
\maketitle
\vspace{-25pt}
\centering
\begin{tcolorbox}[enhanced, width=0.97\linewidth,
  colback=black!3, colframe=black!12, boxrule=0.5pt, arc=6pt,
  left=6pt, right=6pt, top=5pt, bottom=5pt]
\centering
{\small
\makebox[0.25\linewidth]{\textbf{Partial input}}%
\makebox[0.25\linewidth]{\textbf{\textcolor{frozen}{Frozen global model}}}%
\makebox[0.25\linewidth]{\textbf{\textcolor{cargo}{\mymethod{} (ours)}}}%
\makebox[0.25\linewidth]{\textbf{Ground truth}}\par}
\vspace{3pt}
\includegraphics[width=\linewidth]{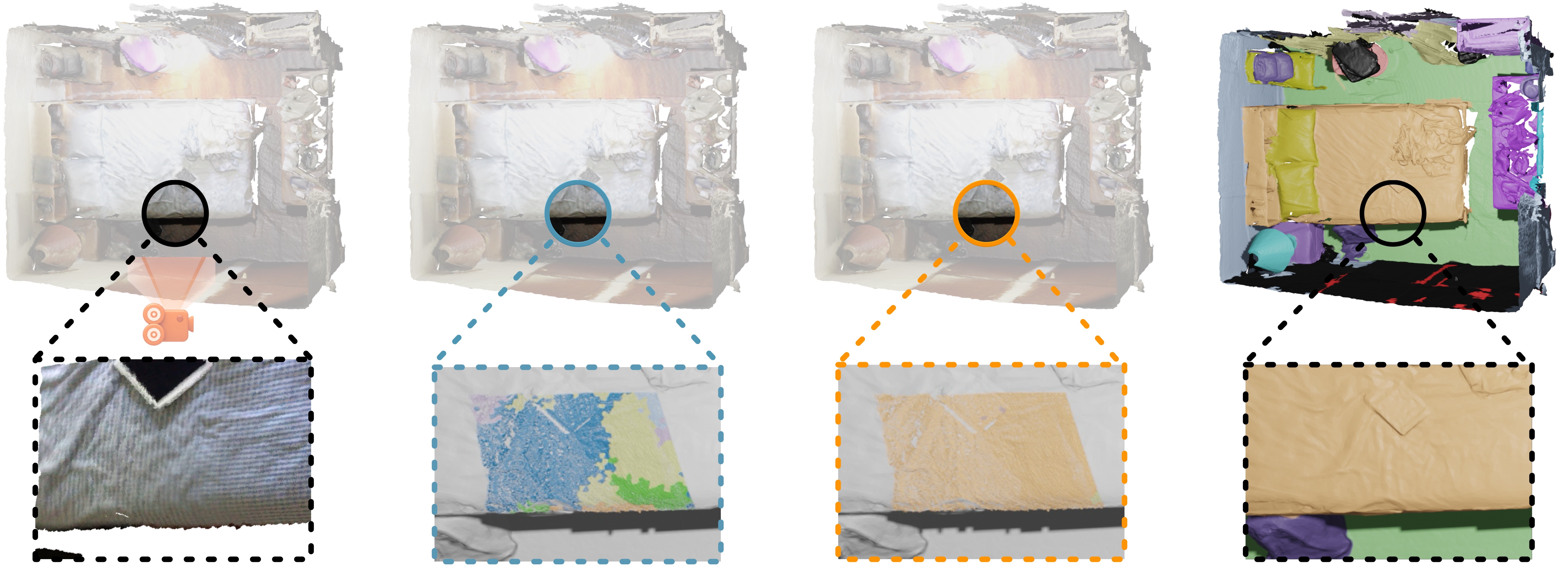}\par
\vspace{3pt}
{\small
\makebox[0.25\linewidth]{Camera frame, limited FoV}%
\makebox[0.25\linewidth]{\chip{wallc}~wall~\textcolor{red!70!black}{\xmark}}%
\makebox[0.25\linewidth]{\chip{bedc}~bed~\textcolor{green!50!black}{\cmark}}%
\makebox[0.25\linewidth]{Full scene, for reference}\par}
\end{tcolorbox}
\vspace{-2pt}
\captionof{figure}{\textbf{Problem formulation: partial semantic segmentation.}
Given a single-view 3D point cloud with limited field of view (FoV), \eg, from an RGB-D frame or a monocular 3D reconstruction (left), our goal is to predict semantic labels for the \textbf{visible} points (circle) without relying on additional scene context.
This setting mirrors semantic scene understanding from onboard cameras, where a model must infer object semantics from partial, incomplete observations in its own camera frame.
Here, the bed is hard to identify: the view is limited, and the bed is white and flat like the surrounding surfaces.
The globally pretrained model~\cite{sonata} misclassifies it as \chip{wallc}~wall, while \mymethod{} recovers \chip{bedc}~bed without labels, pose, or gravity.
\textcolor{gray}{Transparent} points denote unobserved regions, shown only for reference.%
\label{fig:teaser}}
\vspace{1em}
}]
\begin{abstract}
Depth sensors and feed-forward reconstruction now yield 3D point clouds from a single view, but as partial observations in camera coordinates, unlike the complete world-frame scenes that 3D encoders are pretrained on. On ScanNet, a frozen Sonata encoder with a linear probe drops from 72.47 mIoU on full scenes to 2.57 on single camera-coordinate frames, while the same frames in world coordinates still reach 67.45: the collapse stems mainly from the coordinate frame and then from missing geometry. Concerto fails in the same way, and gravity estimated from the image recovers only 41.64.

We propose \mymethod{}, which trains a 1.3M-parameter adapter using only paired partial and global point clouds, keeping the encoder and global probe frozen. One loss aligns each partial-view feature with the global feature of the same point, a second matches its similarities to features sampled from the full scene. At inference, \mymethod{} takes a single camera-coordinate view, with no pose or gravity, and one adapter serves both the ScanNet-20 and ScanNet200 probes.

Without pose or gravity, \mymethod{} reaches 65.46 mIoU on ScanNet-20, and its Gram variant 67.27, above ground-truth gravity alignment (66.49) and on par with the ground-truth-pose oracle (67.45). Given gravity, \mymethod{} reaches 68.27, above the oracle that bounds any correction of the input frame. It outperforms label-supervised PEFT on ScanNet-20 with both encoders and on all three benchmarks with Concerto, and in ScanNet→ScanNet++ transfer it exceeds fully fine-tuned Sonata (63.44 vs. 56.58).

\end{abstract}

\section{Introduction}
\label{sec:intro}
Large-scale 3D representation learning has advanced semantic scene understanding, enabling pretrained point-cloud encoders to transfer to downstream tasks such as semantic/instance segmentation and scene parsing~\citep{pointnet,pointtransformer,pointtransformerv3,mask3d,pointcontrast,maskedscenecontrast,groupcontrast,sonata,concerto}. Most existing models, however, are developed and evaluated on globally reconstructed scenes, where geometry is aggregated across views in a consistent world-coordinate frame~\citep{scannet, scannetpp}. This differs from realistic embodied perception, where an agent often observes only a partial point cloud from its current viewpoint, with incomplete geometry, occlusions, limited FoV, and camera-coordinate inputs (Fig.\,\ref{fig:teaser}), \eg, before a map exists, in reactive control, or after the scene has changed. Many tasks act only on the current view: a robot passing through a doorway or placing an object on a table needs semantics for the door or table in front of it, not a segmentation of the whole room that must be rebuilt whenever objects move. Similar viewpoint gaps have been studied in egocentric and ego-exo video representation learning~\citep{ego4d,egoexo4d,rosetta}, but their effect on pretrained 3D point-cloud representations remains underexplored.

We show that this gap is not merely a consequence of missing geometry, but a severe representation mismatch between globally learned 3D feature spaces and view-constrained partial observations. Using Sonata~\citep{sonata} as a strong representative pretrained encoder, we find that a frozen encoder with a global linear probe reaches $72.47$~mIoU on full ScanNet scenes, but drops to $2.57$~mIoU on single-frame camera-coordinate partial views. Notably, when the same partial views remain expressed in world coordinates, performance remains much higher at $67.45$ mIoU (Sec.~\ref{sec:gap}). This indicates that camera-coordinate partial inputs can lose compatibility with the semantic feature space learned from globally reconstructed scenes. Recent progress in monocular and multi-view 3D reconstruction~\citep{dust3r2024,vggt,pi32024} further motivates this setting, as view-constrained 3D observations are becoming increasingly available from casually captured visual input.

A straightforward solution is to fully fine-tune the encoder and a segmentation head on labelled camera-view point clouds.  This is effective in-domain, but needs a separately adapted model and dense partial-view labels for every target setting, and can hurt cross-dataset generalization.
 Parameter-efficient fine-tuning (PEFT)
reduces the number of updated parameters~\citep{peft,lora,prefix,bitfit}, and
geometry-aware methods such as GEM~\citep{gem} have shown that
lightweight spatial adaptation can be effective for 3D scene segmentation. However, existing PEFT methods rely on semantic supervision in the target domain and have not been applied to partial-view segmentation, where label biases and missing scene context further limit their effectiveness.

To address this, we propose \mymethod{}, a partial-to-global adaptation framework
that maps partial camera-coordinate observations into the semantic
feature space of a frozen globally pretrained model. Our key premise is that
adapting individual features independently is insufficient: a partial
observation should recover not only the global feature associated with the
same physical point, but also its relationships to the broader global
representation.

We therefore derive two complementary forms of supervision entirely from
paired partial/global geometry. First, matched-point feature anchoring aligns
each partial feature with the global feature of the corresponding physical
point. Second, a point-specific relational objective preserves how that
feature relates to reference features sampled from the global scene. This
transfers structure from the complete-scene representation without requiring
semantic annotation of the partial observations. The pretrained encoder and
global segmentation probe remain frozen throughout adaptation. Paired global
geometry is used only during training. At inference, \mymethod{} operates
directly on partial camera-coordinate point clouds and requires neither the
complete scene nor camera pose.

We evaluate \mymethod{} under a partial-view semantic segmentation protocol constructed from standard 3D scene datasets, including ScanNet and ScanNet++~\citep{scannet,scannetpp}. On ScanNet, \mymethod{} reaches $65.46$ mIoU, outperforming adapter-based PEFT~\citep{peft} and GEM~\citep{gem} without %
partial-view semantic labels. 
In cross-dataset transfer from ScanNet to ScanNet++, \mymethod{} reaches $63.44$ mIoU and surpasses the fully fine-tuned ScanNet Sonata baseline, suggesting that aligning to the frozen global representation can preserve transferability better than %
adapting the backbone to a labelled source domain. 
We summarize the contributions as follows:

\begin{itemize}[nosep, leftmargin=2em]
    \item We isolate coordinate-frame mismatch, not missing geometry, as the dominant failure of globally pretrained 3D encoders on partial views ($-67$ vs.\ $-5$ mIoU).
    \item We propose \mymethod{}, which repairs this mismatch from paired partial/global geometry alone, keeping the encoder and probe frozen, requiring no global views or camera pose at inference.
    \item With $1.3$M parameters and no partial-view labels, \mymethod{} outperforms supervised PEFT in-domain~($65.46$~mIoU) on ScanNet-20 and fully fine-tuned Sonata under domain shift ($63.44$ vs.\ $56.58$).
\end{itemize}
\section{Related Work}
\label{rwork}
\textbf{3D representation learning.}
Point-based and transformer-based architectures have driven rapid progress in 3D semantic segmentation \citep{Neshev2025Seg3dsurvey,pointnet,pointtransformer,pointtransformerv3,mask3d}. In parallel, self-supervised pretraining has become the main route to transferable 3D features~\citep{Xiao2023PointRep}. PointContrast introduced correspondence-based pretraining on aligned RGB-D observations\,\citep{pointcontrast}. Masked Scene Contrast and GroupContrast scaled scene-level pretraining and improved semantic grouping\,\citep{maskedscenecontrast,groupcontrast}; and Sonata explicitly addressed geometric shortcut learning to produce stronger frozen representations\,\citep{sonata}. However, these models are trained and evaluated primarily on globally reconstructed scenes. Our focus is the deployment gap that arises when such representations are applied to single-view partial observations.
\\
\textbf{Egocentric and cross-view understanding.}
Cross-view learning has been studied extensively in video through ego/exo
datasets and view-invariant representation
learning~\citep{ego4d,egoexo4d,rosetta}. In contrast to video, where viewpoint
invariance is an established objective, the analogous gap in pretrained 3D
point-cloud representations has received little attention. At the same time,
recent 3D reconstruction systems such as DUSt3R, VGGT, and
PI3~\citep{dust3r2024,vggt,pi32024} make dense partial observations from
monocular capture readily available, making this gap increasingly consequential.
Dense per-point semantic segmentation from a single partial camera-coordinate
cloud, without multi-frame fusion or reconstruction, remains largely unexplored.
\\
\textbf{Adaptation and PEFT for 3D.}
PEFT methods such as adapters\,\citep{peft}, LoRA\,\citep{lora}, prefix
tuning\,\citep{prefix}, and BitFit\,\citep{bitfit} are effective in language
and vision but transfer poorly to point clouds, as they adapt at the per-point
level or inject static global tokens, ignoring local spatial
structure\,\citep{gem}. GEM\,\citep{gem} addresses this with geometry-aware
adapters, approaching full fine-tuning with few trainable parameters, but
operates on fully reconstructed scenes and relies on semantic supervision.

In contrast, \mymethod{} targets a more challenging deployment setting: adapting
a pretrained global 3D encoder to view-constrained point clouds in
camera coordinates, where both scene context and the coordinate frame differ from
pretraining. Rather than relying on semantic labels, \mymethod{} learns a
lightweight translator from paired partial/global geometry alone. Its
self-distillation objective transfers both local and scene-level structure by
matching similarity relations to shared global keys, addressing the combined
effects of coordinate mismatch and limited context in local views.
\\
\textbf{Rotation robustness and online adaptation.}
Rotation-invariant and -equivariant point networks~\citep{vectorneurons,riconv} remove frame dependence by design, but require retraining the backbone and often trade accuracy on canonically oriented data, which forfeits the pretrained encoders we aim to reuse. Test-time adaptation updates a pretrained 3D model on unlabeled target inputs~\citep{mate}, and online 3D perception fuses per-frame features into a growing scene representation~\citep{fusionaware,memadapter}. \mymethod{} is not an alternative to such mapping but can improve the per-frame predictions they fuse. It is trained once from paired geometry and runs per frame without fusion, pose, or test-time optimization.
\\
\textbf{Relational distillation.}
Matching relations between samples rather than individual outputs is established in knowledge distillation~\citep{rkd,spkd}, and similarity distributions over a set of anchors have been used to compress self-supervised models~\citep{compress,seed}. Complete-to-partial distillation has trained LiDAR segmentation and scene-completion networks with semantic labels~\citep{m2skd,scpnet} and self-supervised 4D sequence encoders~\citep{c2p}. We instead adapt a frozen encoder and probe without labels, to a change of input frame that none of these addresses, with keys from a complete scene most of which the student does not observe. Sec.~\ref{sec:why} characterizes what this term constrains beyond pointwise alignment.

\section{Representation Gap Analysis}
\label{sec:gap}
\begin{figure*}[t]
\centering
\setlength{\gw}{0.965\linewidth}%
\footnotesize
\begin{tabular}{@{}c@{\hspace{4pt}}l@{}}
 & \makebox[0.570\gw]{\color{gray}\hrulefill\ \textcolor{black}{\textbf{World} coordinates
     \textcolor{worldc}{($-$5.0 mIoU)}}\ \hrulefill}\hspace{0.0108\gw}%
   \makebox[0.4176\gw]{\color{gray}\hrulefill\ \textcolor{camc}{\textbf{Camera} coordinates
     ($-$67.3 mIoU)}\ \hrulefill} \\[3pt]
\rotatebox[origin=c]{90}{\textbf{Point cloud}} &
  \raisebox{-0.5\height}{\includegraphics[width=\gw]{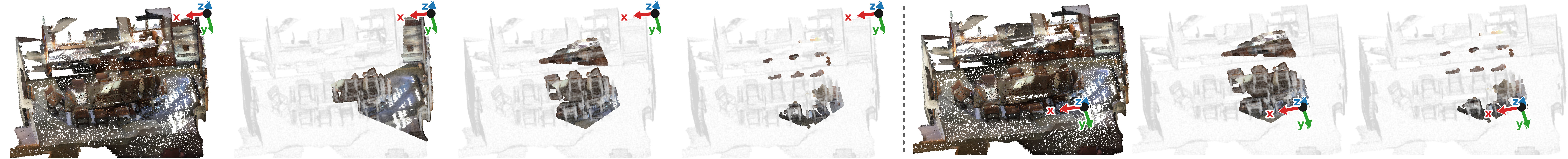}} \\[1pt]
\rotatebox[origin=c]{90}{\textbf{PCA features}} &
  \raisebox{-0.5\height}{\includegraphics[width=\gw]{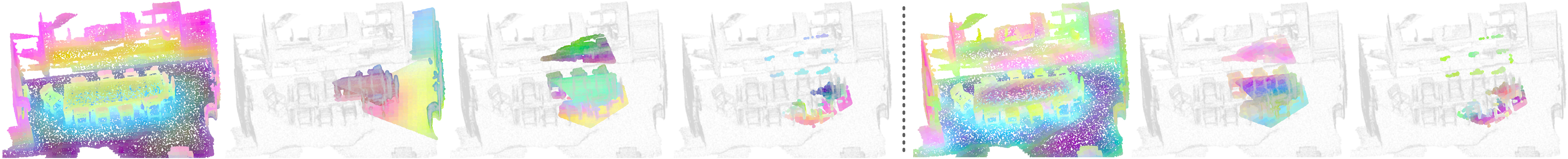}} \\[3pt]
 & \makebox[0.1425\gw]{\gaplab{Global}{\miou{worldc}{\textbf{72.47}}}}%
   \makebox[0.1425\gw]{\gaplab{Room slice}{\miou{worldc}{69.64}}}%
   \makebox[0.1425\gw]{\gaplab{View crop}{\miou{worldc}{69.02}}}%
   \makebox[0.1425\gw]{\gaplab{Single frame}{\miou{worldc}{67.45}}}%
   \hspace{0.0108\gw}%
   \makebox[0.1392\gw]{\gaplab{Global}{\miou{camc}{5.14}}}%
   \makebox[0.1392\gw]{\gaplab{View crop}{\miou{camc}{3.00}}}%
   \makebox[0.1392\gw]{\gaplab{Single frame}{\miou{camc}{\textbf{2.57}}}} \\
\end{tabular}
\caption{\textbf{Two sources of degradation} (frozen Sonata with the global linear probe, ScanNet; mIoU under each input).
\emph{Left:} in world coordinates, removing geometry step by step, from the complete scene down to a single frame, lowers mIoU by only 5.0 points.
\emph{Right:} expressing the same inputs in camera coordinates collapses performance, even for the complete scene (72.47 $\rightarrow$ 5.14), and the PCA features (bottom row) change visibly.
The coordinate frame is therefore the dominant factor, and missing geometry further compounds the gap.
Triads mark the origin and the $x$, $y$, $z$ axes (red, green, blue) of each input frame.}
\label{fig:gap_vis}
\end{figure*}
Before introducing \mymethod{}, we quantify the mismatch between globally pretrained 3D encoders
and view-constrained partial observations. The goal is to isolate \emph{why} the standard
frozen-transfer recipe fails, and %
what the adaptation module needs to correct.
\\
\textbf{Evaluation conditions.}
We construct four input conditions that progressively increase deployment realism and evaluate them in world and camera coordinate frames~(Fig.\,\ref{fig:gap_vis}). \textbf{Global} is the scene-level point cloud accumulated from all frames, corresponding to the conventional training and evaluation setting for 3D scene understanding. \textbf{Room Slice} retains one sixth of the global point cloud, obtained by partitioning the scene into six slices from the room center. It introduces coarse geometric incompleteness while preserving the world coordinate frame. \textbf{View Crop} is a spatial crop of the global point
cloud centered at a single camera viewpoint, retaining only points within the region a single frame would observe while keeping full geometric fidelity within the crop. \textbf{Single Frame} contains the 3D points back-projected from one RGB-D observation, reflecting realistic deployment with limited field of view, missing geometry, and depth noise.

For the \textbf{World} variants, all points are represented in the global world coordinate system. For the \textbf{Camera} variants, points are instead represented in a camera coordinate system. Only the diagnostic {Camera|Global} variant (Fig.~\ref{fig:gap_vis}, 5\textsuperscript{th}~col.) expresses the whole scene in the frame of its first image.  {View Crop} and {Single Frame}~(6\textsuperscript{th} and 7\textsuperscript{th}~col.) use the frame of their own camera, as in all main experiments.
\\
\textbf{Two separable sources of degradation.}
Table~\ref{tab:gap} reveals an asymmetry. Within world coordinates, performance decreases gradually as inputs become more geometrically incomplete: from $72.47$ mIoU on Global down to $67.45$ mIoU on Single Frame, a drop of about 5 points across four conditions spanning the full range from complete to single-frame reconstruction. (This $67.45$ mIoU value uses ground-truth camera pose to transform Single Frame inputs into world coordinates, an oracle reference not available to \mymethod{} at inference.) The pretrained representation is more robust to missing geometry when the coordinate frame is preserved. The coordinate frame tells a completely different story. Expressing the full Global scene in camera coordinates, with no geometry removed at all, already collapses mIoU from $72.47$ to $5.14$. Adding single-frame partiality brings it to $2.57$. The coordinate frame alone, applied to a geometrically complete scene, causes a $67$-point drop. Geometry reduction, by contrast, causes only a $5$-point drop. Within the camera-coordinate setting, geometry reduction continues to degrade performance (5.14 → 2.57 mIoU), confirming that both factors contribute, with the coordinate frame being the larger and qualitatively distinct effect.
\\
\textbf{Why does the coordinate frame matter?}
The Camera|Global condition applies a single rigid transform to an otherwise unchanged complete scene. The 67-point drop therefore cannot be attributed to missing scene content. The pretrained encoder often relies on world-frame priors (floorbelow,
gravity-aligned up-axis, height above the floor) that are disrupted under camera coordinates. Standard augmentations %
could not cover all possible variations introduced by camera frames, leaving this gap unaddressed.
\\
\textbf{Can training-free canonicalization solve the gap?}
The strong coordinate-frame effect raises the question of whether the problem
can be addressed without learning. We therefore evaluate two training-free
canonicalization strategies. PCA-based axis alignment lifts the frozen
model from 2.57 to only 8.38 mIoU. Gravity alignment, with pitch and roll
estimated from the RGB image by Perspective
Fields~\citep{persp-fields}, reaches 41.64 mIoU.

With ground-truth gravity the frozen model reaches 66.49 mIoU: on ScanNet, gravity alone recovers nearly all of the frame gap. This relies on a property of the benchmark rather than of the task: ScanNet scenes are upright, and the encoder was pretrained on gravity-aligned reconstructions. It also requires an accurate gravity sensor aligned with the camera: mIoU holds to $\sigma=5^\circ$ ($66.21$) and drops to $61.52$ at $10^\circ$ (Appendix~\ref{app:gravity}), and image-based estimates give $41.64$.  Even when gravity is exact, it corrects only the frame: the frozen model stays below complete scenes ($66.49$ vs. $72.47$), because no rotation recovers what the partial view does not contain.

\mymethod{} needs neither pose nor gravity. From the camera-frame input alone it reaches $67.27$ mIoU with the regression loss ($65.46$ with KL), above ground-truth gravity alignment ($66.49$) and on par with the ground-truth pose ($67.45$). Given gravity, it reaches $68.27$, above that pose oracle, which bounds any correction of the input frame. The remaining gain therefore is consistent with scene context: the global targets were computed on the complete scene, so aligning to them transfers context that a close-range view lacks, which disambiguates visually similar planar regions.

\begin{table}[t]
\centering
\caption{\textbf{Representation gap on ScanNet} (frozen Sonata and global probe).
The frame costs 67.3 mIoU, missing geometry 5.0. Only ground-truth gravity recovers most of it without training. CARGO does so without pose or gravity.}
\label{tab:gap}
\small
\setlength{\tabcolsep}{3.1pt}
\begin{tabular}{llccc}
\toprule
Frame & Input & allAcc & mAcc & mIoU \\
\midrule
\multirow{4}{*}{\textcolor{worldc}{\textbf{World}}}
 & Global        & 89.73 & 83.27 & \textbf{72.47} \\
 & Room Slice    & 88.48 & 81.01 & 69.64 \\
 & View Crop     & 87.24 & 80.79 & 69.02 \\
 & Single Frame  & 87.36 & 78.98 & 67.45 \\
\midrule
\multirow{6}{*}{\textcolor{camc}{\textbf{Camera}}}
 & Global        & 26.42 &  9.56 &  5.14 \\
 & View Crop     & 19.92 &  7.52 &  3.00 \\
 & Single Frame  & 18.30 &  6.94 &  2.57 \\
 & \quad + PCA align              & 28.39 & 14.70 &  8.38 \\
 & \quad + Gravity (est.)~\citep{persp-fields} & 70.85 & 54.74 & 41.64 \\
 & \quad + Gravity (GT)  & 86.90 & 78.28 & 66.49 \\
\midrule
\ours & KL (no gravity)               & 85.93 & 80.50 & 65.46 \\
\ours & Gram (no gravity)         & 87.62 & 77.75 & 67.27 \\
\ours \multirow{-3}{*}{\textcolor{cargo}{\textbf{+\,\mymethod{}}}}
 & KL + gravity (GT) & \textbf{88.10} & \textbf{82.41} & \textbf{68.27} \\
\bottomrule
\end{tabular}
\end{table}

\section{Methodology}
\label{method}

\begin{figure*}[t]
\centering
\includegraphics[width=0.9\linewidth]{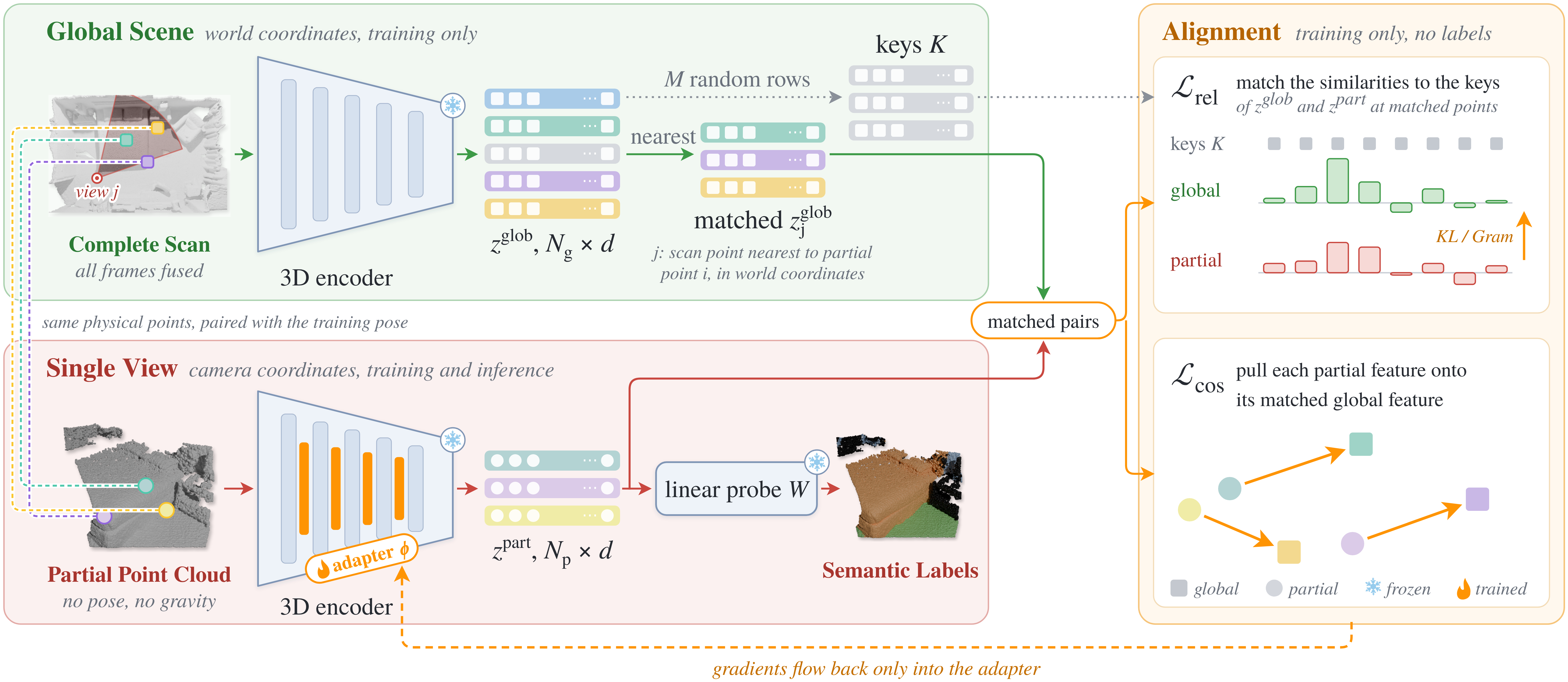}
\caption{\textbf{\mymethod{} architecture.} A global scene (world coordinates) and a partial view (camera coordinates) share one frozen 3D encoder (Sonata or Concerto). Only the adapter $\phi$ is trained. Markers link the same physical points. At matched points, $\mathcal{L}_\text{rel}$ (Eqs.\,\ref{eq:kl},\,\ref{eq:gram}) aligns their similarities to random keys $K$, and $\mathcal{L}_\text{cos}$ (Eq.\,\ref{eq:cos}) aligns the features directly. Only the bottom branch runs at inference.}
\label{fig:model}
\end{figure*}

Let $f$ be a frozen 3D encoder pretrained on globally reconstructed scenes, producing
per-point features in $\mathbb{R}^d$, and let $W \in \mathbb{R}^{C \times d}$ be the frozen
linear probe trained on world-coordinate global scenes for semantic segmentation.

Our goal is to adapt features extracted from partial camera-coordinate observations so that they become compatible with the global feature space learned by $f$ and the frozen classifier $W$, without updating the backbone or requiring semantic labels for the target partial views. 
Adaptation is driven entirely by correspondences to the pretrained global feature space rather than by class-specific supervision. At inference, the model receives only a partial camera-coordinate point cloud and requires neither the complete scene nor camera pose.

Unlike full fine-tuning or label-supervised PEFT, \mymethod{}'s adaptation is self-supervised, like the encoder it adapts, and independent of the label set. Unlike pointwise regression, it also constrains how each feature relates to the rest of the scene.

\subsection{\mymethod{}}
\label{sec:overview}

We design $\phi$ as a lightweight geometry-aware adapter model inserted into the frozen backbone\,(Fig.\,\ref{fig:model}). It refines per-point features through a residual bottleneck whose middle layer aggregates over a local 3D voxel neighbourhood:
\begin{equation}
    \phi(x) = x + \sigma\left(\sum_{i \in \mathcal{N}} W^{i}_{\text{loc}}(x W_{\text{down}})\right) W_{\text{up}},
\end{equation}
where $W_{\text{down}} \in \mathbb{R}^{d \times r}, W_{\text{up}} \in \mathbb{R}^{r \times d}$, and $r \ll d$. The local neighbourhood $\mathcal{N}$ is defined in voxel space, allowing the adapter to aggregate spatially nearby information while remaining parameter-efficient. We adopt a GEM-style spatial adapter~\citep{gem} because
its local-aggregation prior is well matched to correcting the coordinate-frame errors identified in Sec.~\ref{sec:gap}. We
deliberately do \emph{not} include a global-attention module: scene-level information is
supplied through the loss, rather than the architecture and Sec.~\ref{sec:why} analyzes how.\\
\textbf{Notation.} $z^{\text{part}}_i$ is the adapted feature of partial-view point $i$, and $z^{\text{glob}}_j$ the frozen feature of global-scene point $j$. Matched pairs $(i,j)\in\mathcal{A}$ come from nearest neighbors in world coordinates (Appendix~\ref{app:dataset_preprocessing}), the only use of poses.

\subsection{Training Objective}
\label{sec:loss}
\textbf{Local anchor: cosine alignment at matched points.} For each matched pair \((i,j) \in \mathcal{A}\), we align the final partial-view feature
\(z^{\text{part}}_i\) with its corresponding global feature \(z^{\text{glob}}_j\):
\begin{equation}
\label{eq:cos}
\mathcal{L}_{\text{cos}} =
\frac{1}{|\mathcal{A}|}
\sum_{(i,j)\in \mathcal{A}}
\left(
1 -
\cos\left(
z^{\text{part}}_i,
z^{\text{glob}}_j
\right)
\right).
\end{equation}
This term provides a direct local alignment signal. It encourages each adapted partial-view feature to match the global representation of the same physical point, which is already compatible with the frozen linear probe.\\
\textbf{Relational term: point-specific similarity supervision.} Let $K \in \mathbb{R}^{M \times d}$ be the frozen global features of $M$ points drawn uniformly at random from the global scene. A point's similarities to these keys describe how it relates to the rest of the scene. Both branches compute similarity logits over $K$:
\begin{equation}
s^{\text{part}}_i = z^{\text{part}}_i K^\top / \tau_s, \qquad
s^{\text{glob}}_j = z^{\text{glob}}_j K^\top / \tau_t,
\end{equation}
with teacher temperature $\tau_t < \tau_s$ so that the teacher is sharper, following DINO~\citep{dino}. %
We consider two instantiations of the relational term, which differ only in how the two similarity patterns are compared. \emph{KL} compares the teacher and student distributions over the keys:
\begin{equation}
\label{eq:kl}
\mathcal{L}_{\text{KL}} = \frac{1}{|\mathcal{A}|} \sum_{(i,j) \in \mathcal{A}}
D_{\text{KL}}\!\left(\text{softmax}(s^{\text{glob}}_j) \,\|\, \text{softmax}(s^{\text{part}}_i)\right).
\end{equation}
\emph{Gram} regresses the similarities directly, matching the sampled rows of the Gram matrix of partial and global features:
\begin{equation}
\label{eq:gram}
\mathcal{L}_{\text{Gram}} = \frac{1}{|\mathcal{A}|\,M} \sum_{(i,j) \in \mathcal{A}} \big\| (z^{\text{part}}_i - z^{\text{glob}}_j)\, K^\top \big\|_2^2 .
\end{equation}

\textbf{Total objective.} We minimize $\mathcal{L} = \mathcal{L}_{\text{cos}} + \mathcal{L}_{\text{rel}}$, with $\mathcal{L}_{\text{rel}} \in \{\mathcal{L}_{\text{KL}}, \mathcal{L}_{\text{Gram}}\}$ used one at a time, and call the resulting models \mymethod{}-KL and \mymethod{}-Gram. Unless stated otherwise, \mymethod{} denotes \mymethod{}-KL. %
Directly regressing the sampled similarities (Gram) is an equally valid
parameterization of the same constraint. On ScanNet-20 it attains higher mIoU and
allAcc, while KL attains higher mAcc, on both backbones (Tab.~\ref{tab:partial_view_all}); on ScanNet++ and
ScanNet200 the ranking varies (Appendix~\ref{app:relational}). The contribution is therefore the
point-specific relational supervision, not a particular divergence.

Because keys are random samples across the
scene, $\text{softmax}(s^{\text{glob}}_j)$ is a compact signature of how
$z^{\text{glob}}_j$ relates to the rest of the scene. Matching it forces the
student to reproduce that relational signature from partial observation.

\subsection{Why the two losses are complementary}
\label{sec:why}

The cosine loss anchors each partial feature to its global target but says nothing about how features relate to each other. The relational loss fills this gap: by matching similarity patterns over shared keys, it constrains inter-feature structure rather than just absolute positions. Tab.~\ref{tab:ablations}\,(A1) shows the two provide complementary gains.
\\
\textbf{Why random keys?} An alternative is to sample keys from the anchor's neighbourhood (e.g., k-NN in world coordinates), but this is counterproductive: spatially adjacent points lie on similar surfaces and produce highly correlated inner products with the anchor. Random sampling avoids this by drawing keys broadly across the scene, while remaining O(M) and trivially parallelizable, in contrast to alternatives such as farthest-point sampling whose cost scales poorly with scene size. Tab.~\ref{tab:ablations}\,(A2) confirms the predicted drop when $k$-NN keys replace random
keys.

\section{Experiments}
\label{sec:setup}
We report results on two
pretrained encoders, Sonata~\citep{sonata} and Concerto~\citep{concerto}, and
use Sonata for ablations. An frozen-transfer comparison
with PTv3~\citep{pointtransformerv3} is in Appendix~\ref{app:encoder}.
\\
\textbf{Datasets.} We use ScanNet~\citep{scannet}, ScanNet200~\citep{rozenberszki2022language},
and ScanNet++~\citep{scannetpp}. ScanNet contains $1513$ RGB-D scans with $20$ and
$200$ class point-level annotations. Each frame is back-projected using its %
captured camera pose, so the
evaluation set spans ScanNet's natural pose distribution (Appendix~\ref{app:pose}).
We sample every $100^{\text{th}}$ frame for training~($\sim$0.3\,FPS, $\approx$19K frames) and every $30^{\text{th}}$ frame for evaluation
($\sim$1\,FPS, $\approx$18K frames). ScanNet++ provides high-fidelity scans with $100$
classes. We sample every $60^{\text{th}}$ frame for evaluation ($\sim$1\,FPS,~$\approx$7K frames). All methods are trained and evaluated on the same subsets for fair comparison. Full statistics are in Appendix~\ref{sec:appendix}.
\\
\textbf{Baselines.}
We compare against methods with increasing label and parameter cost. We evaluate two PEFT baselines: \textit{Adapter}~\citep{peft}, which inserts small trainable modules before the classifier, and GEM~\citep{gem}, a geometry-aware PEFT method that adds spatial and context adapters to the frozen encoder and trains them with supervised cross-entropy. We also include \linearprobing{}, which trains a linear classifier on frozen partial-view features, the standard protocol for evaluating pretrained 3D representations~\citep{sonata}. \late{}, additionally unfreezes the final encoder blocks and \full, updates the entire encoder. All baselines require semantic labels, while \mymethod{} does not.
\\
\textbf{Metrics.} We evaluate all methods on Intersection-over-Union\,(IoU) between predicted and ground truth masks. Consistent with common practices in the field we also report mean and overall Accuracy. 
\begin{table*}[t]
\centering
\caption{\textbf{Semantic segmentation under partial-view adaptation} with two pretrained 3D backbones.
\textbf{Bold} and \underline{underline} mark the best and second-best results per backbone.
\mymethod{} uses no partial-view labels (\nolab); all baselines do (\lab).}
\label{tab:partial_view_all}
\small
\setlength{\tabcolsep}{4.2pt}
\begin{tabular}{lcc ccc ccc ccc}
\toprule
 & & & \multicolumn{3}{c}{ScanNet-20} & \multicolumn{3}{c}{ScanNet++} & \multicolumn{3}{c}{ScanNet200} \\
\cmidrule(lr){4-6}\cmidrule(lr){7-9}\cmidrule(l){10-12}
Method & Labels & Params & mIoU$\uparrow$ & mAcc$\uparrow$ & allAcc$\uparrow$ & mIoU$\uparrow$ & mAcc$\uparrow$ & allAcc$\uparrow$ & mIoU$\uparrow$ & mAcc$\uparrow$ & allAcc$\uparrow$ \\
\midrule
\multicolumn{12}{l}{\textit{Sonata}~\citep{sonata} \venue{arxiv'25}} \\
\base{Frozen}                 & \base{--} & \base{--} & \base{2.57} & \base{6.94} & \base{18.30} & \base{0.69} & \base{1.92} & \base{9.78} & \base{0.14} & \base{0.37} & \base{3.11} \\
\base{Full + linear}$^\dagger$ & \lab & \base{108.5M} & \base{69.26} & \base{77.11} & \base{87.88} & \base{35.99} & \base{44.75} & \base{85.95} & \base{24.58} & \base{32.13} & \base{77.86} \\
+ Linear                      & \lab & $<$0.1M & 36.17 & 52.42 & 68.64 & 15.73 & 26.11 & 67.66 &  8.13 & 12.98 & 61.43 \\
+ Late blocks + linear        & \lab & 21.0M   & 48.28 & 59.13 & 76.86 & 17.77 & 25.33 & 72.98 & 11.43 & 16.56 & 63.55 \\
+ Adapter~\citep{peft} \venue{ICML'19} & \lab & 6.0M & 41.49 & 65.68 & 74.22 & 13.96 & 28.97 & 65.58 & 10.15 & 16.17 & 64.49 \\
+ GEM~\citep{gem} \venue{NeurIPS'25} & \lab & 1.8M & 59.81 & 73.92 & 83.90 & \textbf{33.27} & \textbf{42.57} & \textbf{84.22} & \textbf{21.97} & \textbf{32.52} & \sbest{75.45} \\
\ours + \mymethod{}-KL (ours)   & \nolab & 1.3M & \sbest{65.46} & \textbf{80.50} & \sbest{85.93} & \sbest{27.38} & \sbest{39.18} & \sbest{81.43} & 19.25 & 27.73 & 74.19 \\
\ours + \mymethod{}-Gram (ours) & \nolab & 1.3M & \textbf{67.27} & \sbest{77.75} & \textbf{87.62} & 21.17 & 25.97 & 80.84 & \sbest{21.10} & \sbest{31.42} & \textbf{76.24} \\ 
\midrule
\multicolumn{12}{l}{\textit{Concerto}~\citep{concerto} \venue{NeurIPS'25}} \\
\base{Frozen}                 & \base{--} & \base{--} & \base{4.97} & \base{10.83} & \base{21.55} & \base{1.44} & \base{4.26} & \base{17.46} & \base{0.37} & \base{1.64} & \base{5.20} \\
+ Linear                      & \lab & $<$0.1M & 39.53 & 56.57 & 71.10 & 18.44 & 29.26 & 70.23 & 12.68 & 19.94 & 65.89 \\
+ GEM~\citep{gem} \venue{NeurIPS'25} & \lab & 1.8M & 69.90 & 80.96 & 88.64 & 32.86 & \sbest{45.28} & 84.12 & 25.06 & 35.10 & \sbest{77.83} \\
\ours + \mymethod{}-KL (ours)   & \nolab & 1.3M & \sbest{71.67} & \textbf{87.77} & \sbest{89.96} & \textbf{35.27} & \textbf{48.49} & \sbest{85.57} & \sbest{26.73} & \textbf{50.19} & 76.12 \\
\ours + \mymethod{}-Gram (ours) & \nolab & 1.3M & \textbf{74.14} & \sbest{85.26} & \textbf{90.66} & \sbest{32.99} & 40.10 & \textbf{85.94} & \textbf{28.73} & \sbest{42.32} & \textbf{78.94} \\
\bottomrule
\end{tabular}

\vspace{2pt}
\begin{minipage}{\linewidth}\footnotesize
$^\dagger$Fully supervised full-backbone fine-tuning, shown as an upper bound and excluded from the ranking.
\end{minipage}
\end{table*}

\subsection{Results}
\label{sec:results}

\textbf{In-domain adaptation.} On ScanNet, both variants outperform label-supervised PEFT on both encoders without any labels in the adaptation (Tab.~\ref{tab:partial_view_all}). With Sonata, \mymethod{}-KL reaches $65.46$ mIoU and \mymethod{}-Gram $67.27$, versus $59.81$ for GEM~\citep{gem}. With Concerto, they reach $71.67$ and $74.14$, versus $69.90$. Linear probing and adapter-based PEFT~\citep{peft} recover only partially ($36.17$--$41.49$ mIoU). The two variants trade off as noted in Sec.~\ref{sec:loss}: \mymethod{}-Gram is higher in mIoU
and allAcc, \mymethod{}-KL in mAcc. Both exceed GEM in mAcc (KL: 80.50 vs.\ 73.92 on
Sonata, 87.77 vs.\ 80.96 on Concerto). With only $1.2\%$ of the trainable parameters, they close $94.3\%$ (KL) and $97.0\%$ (Gram) of the gap between the frozen model and full fine-tuning on Sonata ($69.26$ mIoU). On the finer-grained taxonomies, both variants outperform GEM with Concerto on ScanNet200 ($26.73$ KL and
$28.73$ Gram vs.\ $25.06$) and on ScanNet++ ($35.27$ KL and $32.99$ Gram vs.\ $32.86$). With Sonata, both variants trail GEM on ScanNet++ and ScanNet200 (\mymethod{}-KL:
$27.38$ and $19.25$, \mymethod{}-Gram: $21.17$ and $21.10$, GEM: $33.27$ and $21.97$)
but remain above all other baselines except full fine-tuning. Because \mymethod{} reuses the frozen global probe, whereas GEM retrains its head on partial-view labels, its accuracy is bounded by that probe, which likely limits it on the finer 100- and 200-class taxonomies. Qualitative results are shown in Fig.~\ref{fig:qual} and per-class results in Appendix~\ref{app:perclass}. 
\\
\textbf{Backbone generality.} Sonata~\citep{sonata} and Concerto~\citep{concerto} differ substantially in pretraining: point-only self-supervision versus joint 2D-3D self-supervision. Yet both collapse on partial camera-coordinate input ($2.57$ and $4.97$ mIoU, Tab.~\ref{tab:partial_view_all}), indicating that the representation gap identified in Sec. \ref{sec:gap} is a property of global pretraining rather than of a particular encoder. \mymethod{} uses the same adapter and objectives for both and requires only dense per-point features, with no encoder-specific components. That the stronger Concerto representation yields consistent gains over supervised GEM across all three benchmarks suggests \mymethod{} benefits directly from improvements in the pretrained encoder.\\
\textbf{Cross-domain evaluation.} To assess robustness to distribution shifts, we train \mymethod{} and all baselines exclusively on ScanNet, %
using only ScanNet partial/global geometry pairs for \mymethod{}'s self-distillation, and evaluate on ScanNet++ %
without further adaptation. No ScanNet++ labels are used and no adaptation is performed on ScanNet++. Sonata's self-supervised pretraining includes ScanNet++ scenes without labels, which holds for every compared method. We map predictions to the 20 ScanNet classes and report mIoU over the 17 that occur in ScanNet++ (mapping in Appendix~\ref{app:mapping}). As shown in Tab.~\ref{tab:cros}, \mymethod{} transfers better than every baseline. Full fine-tuning is commonly treated as an upper bound, and on ScanNet it is ($69.26$ vs.\ $65.46$ mIoU for \mymethod{}). That bound, however, is local to the source domain: it costs 108.5M updated parameters and dense partial-view labels per setting, and on ScanNet++ the order reverses ($56.58$ vs.\ $63.44$). \mymethod{} keeps the frozen backbone, and with it the generality of the pretrained representation.

\begin{table}[t]
\centering
\caption{\textbf{Cross-domain evaluation} (Sonata). Trained on ScanNet, evaluated on ScanNet++ without adaptation; mIoU over the 17 shared ScanNet-20 classes.}
\label{tab:cros}
\small
\setlength{\tabcolsep}{5pt}
\begin{tabular}{lcccc}
\toprule
Method & Labels & mIoU$\uparrow$ & mAcc$\uparrow$ & allAcc$\uparrow$ \\
\midrule
+ Linear                 & \lab & 31.07 & 35.10 & 72.32 \\
+ Late blocks + linear   & \lab & 40.17 & 42.64 & 78.78 \\
+ Full + linear          & \lab & \sbest{56.58} & 55.10 & \sbest{85.57} \\
+ Adapter~\citep{peft} & \lab & 36.11 & 47.95 & 74.93 \\
+ GEM~\citep{gem}         & \lab & 54.23 & \sbest{56.17} & 84.19 \\
\ours + \mymethod{}-KL (ours) & \nolab & \textbf{63.44} & \textbf{66.64} & \textbf{89.22} \\
\bottomrule
\end{tabular}
\end{table}

\begin{figure}[b]
\centering
\begin{tikzpicture}
\begin{axis}[
  set layers, width=0.85\linewidth, height=5.0cm,
  ybar, bar width=10pt, ymin=0, ymax=79, enlarge x limits=0.13,
  symbolic x coords={sd,d,l,u,su}, xtick=data,
  xticklabels={{\strut steep down\\[-1pt]\textcolor{black!50}{5{,}567}},
               {\strut down\\[-1pt]\textcolor{black!50}{9{,}875}},
               {\strut level\\[-1pt]\textcolor{black!50}{2{,}183}},
               {\strut up\\[-1pt]\textcolor{black!50}{166}},
               {\strut steep up\\[-1pt]\textcolor{black!50}{30}}},
  xticklabel style={font=\scriptsize, align=center},
  ylabel={mIoU}, ylabel style={font=\scriptsize, yshift=-5pt},
  yticklabel style={font=\scriptsize}, ytick={0,20,40,60},
  ymajorgrids, grid style={black!10}, axis line style={black!40}, tick style={draw=none},
  axis x line*=bottom, axis y line*=left,
  legend style={font=\scriptsize, draw=none, fill=none, at={(0.985,0.80)}, anchor=north east, legend columns=1,
                /tikz/every even column/.append style={column sep=6pt}},
  legend image code/.code={\fill[#1] (0cm,-0.07cm) rectangle (0.2cm,0.11cm);},
]
\begin{pgfonlayer}{axis background}
  \fill[black!5] ([xshift=-19pt]axis cs:u,0) rectangle ([xshift=19pt]axis cs:su,79);
\end{pgfonlayer}
\node[font=\tiny, text=black!50, anchor=north] at ([xshift=12pt]axis cs:u,78) {1.1\% of frames};
\addplot[fill=black!30, draw=none] coordinates {(sd,52.52) (d,60.89) (l,52.45) (u,24.55) (su,18.95)};
\addplot[fill=cargo, draw=none] coordinates {(sd,61.31) (d,65.82) (l,61.44) (u,34.63) (su,18.38)};
\node[font=\scriptsize\bfseries, text=cargo!80!black, anchor=south] at (axis cs:sd,61.31) {+8.8};
\node[font=\scriptsize\bfseries, text=cargo!80!black, anchor=south] at (axis cs:d,65.82) {+4.9};
\node[font=\scriptsize\bfseries, text=cargo!80!black, anchor=south] at (axis cs:l,61.44) {+9.0};
\node[font=\scriptsize\bfseries, text=cargo!80!black, anchor=south] at (axis cs:u,34.63) {+10.1};
\node[font=\scriptsize\bfseries, text=cargo!80!black, anchor=south] at (axis cs:su,18.95) {$-$0.6};
\legend{GEM~\citep{gem}, \mymethod{}}
\end{axis}
\end{tikzpicture}
\caption{\textbf{mIoU by camera pitch} (ScanNet-20 partial views, Sonata), with the number of frames per bin and \mymethod{}'s gain over GEM. Bins: steep down ($<-30^\circ$), down ($-30^\circ$ to $-10^\circ$), level ($\pm10^\circ$), up ($10^\circ$ to $30^\circ$), steep up ($>30^\circ$). The shaded upward bins hold only 1.1\% of the frames.}
\label{fig:pitch}
\end{figure}

\subsection{Analysis Experiments}
\label{sec:analysis}

\definecolor{cWall}{RGB}{174,199,232}
\definecolor{cFloor}{RGB}{152,223,138}
\definecolor{cBed}{RGB}{255,187,120}
\definecolor{cChair}{RGB}{188,189,34}
\definecolor{cTable}{RGB}{255,152,150}
\definecolor{cDoor}{RGB}{214,39,40}
\definecolor{cWindow}{RGB}{197,176,213}
\definecolor{cBookshelf}{RGB}{148,103,189}
\definecolor{cPicture}{RGB}{196,156,148}
\definecolor{cDesk}{RGB}{247,182,210}
\definecolor{cToilet}{RGB}{44,160,44}
\providecommand{\sw}[1]{\tikz\fill[#1] (0,0) circle (0.8ex);}
\begin{figure*}[b]
\centering
\setlength{\tabcolsep}{0pt}
\newlength{\qh}\setlength{\qh}{0.238\linewidth}%
\newlength{\qa}\setlength{\qa}{\dimexpr\qh*2002/1000\relax}%
\newlength{\qb}\setlength{\qb}{\dimexpr(\qh-3pt)*1988/1000\relax}%
\begin{tabular}{@{}c@{\hspace{0.02\linewidth}}c@{\hspace{3pt}}c@{}}
{\scriptsize
 \makebox[0.25\qa]{\textbf{(a)}~Input}\makebox[0.25\qa]{GEM~\citep{gem}}%
 \makebox[0.25\qa]{\textcolor{cargo}{\textbf{\mymethod{} (ours)}}}\makebox[0.25\qa]{GT}} &
 &
{\scriptsize
 \makebox[0.333\qb]{\textbf{(b)}~Input}\makebox[0.333\qb]{GEM~\citep{gem}}%
 \makebox[0.333\qb]{\textcolor{cargo}{\textbf{\mymethod{} (ours)}}}} \\[1pt]
\raisebox{-\height}{%
  \includegraphics[height=\qh, trim=0 0 670.3bp 0, clip]{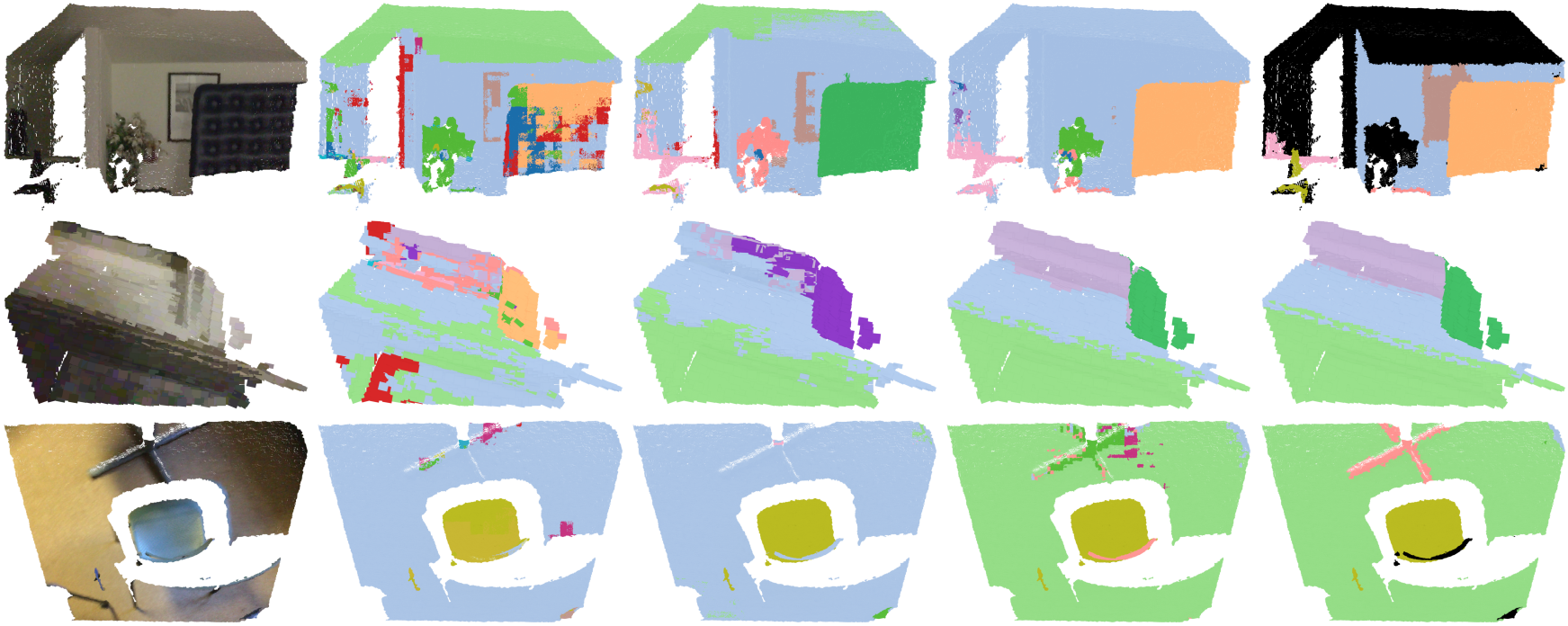}%
  \includegraphics[height=\qh, trim=334.4bp 0 0 0, clip]{figures/grid_2.pdf}} &
\raisebox{-\height}{\begin{tabular}[t]{@{}c@{}}
  \rotatebox[origin=c]{90}{\scriptsize\textit{Monocular}}\rule[-0.09\qb]{0pt}{0.18\qb}\\[3pt]
  \rotatebox[origin=c]{90}{\scriptsize\textit{Robot stereo}}\rule[-0.16\qb]{0pt}{0.323\qb}
\end{tabular}} &
\raisebox{-\height}{\begin{minipage}[t]{\qb}
  \includegraphics[width=\qb]{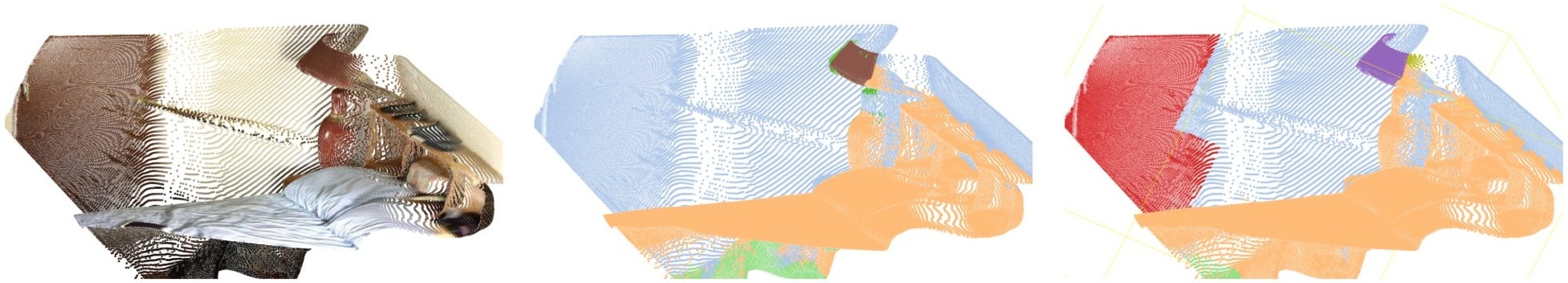}\par\vspace{3pt}
  \includegraphics[width=\qb]{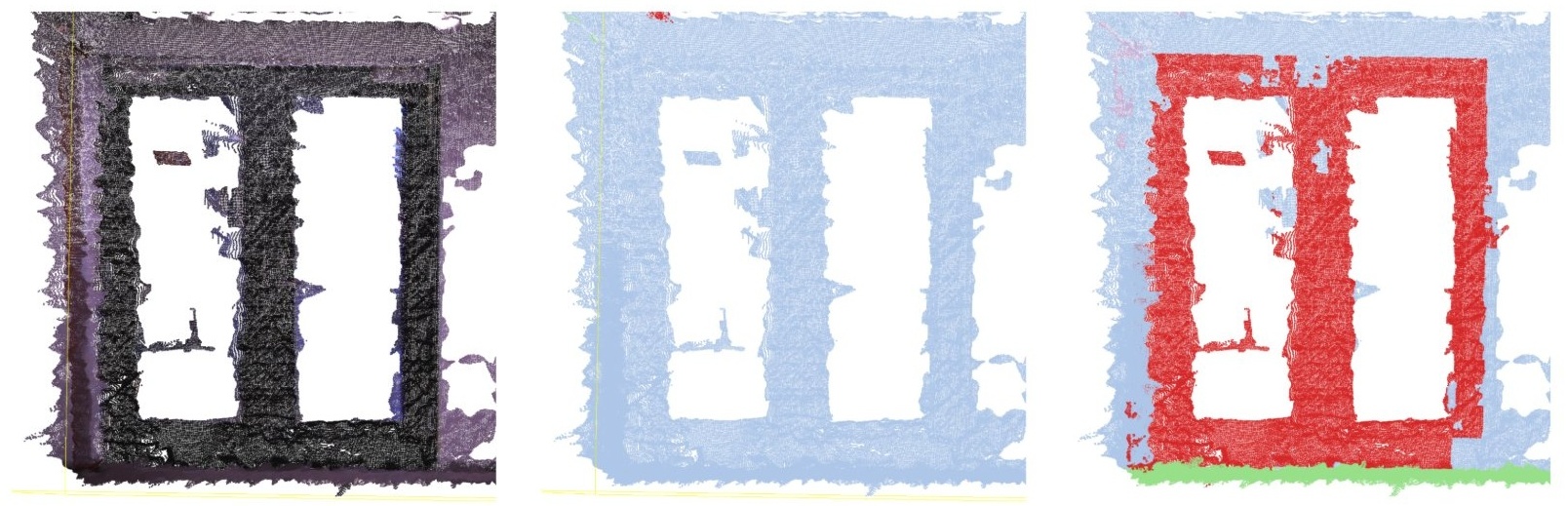}
\end{minipage}} \\
\end{tabular}

\vspace{2pt}
{\scriptsize
\sw{cWall}~wall\quad \sw{cFloor}~floor\quad \sw{cBed}~bed\quad \sw{cChair}~chair\quad
\sw{cTable}~table\quad \sw{cDoor}~door\quad \sw{cWindow}~window\quad \sw{cBookshelf}~shelf\quad
\sw{cPicture}~picture\quad \sw{cDesk}~desk\quad \sw{cToilet}~toilet\par}
\caption{\textbf{Qualitative results.}
\textbf{(a)}~On ScanNet partial views, GEM confuses similar planar surfaces: the mattress (\textit{1st row}) is labeled toilet and the floor (\textit{3rd row}) wall. \mymethod{} keeps the object semantics and labels unannotated points (black) plausibly.
\textbf{(b)}~In the wild, on single-image reconstructions~\citep{vggt} (top) and robot stereo in unseen buildings (bottom), \mymethod{} recovers the \textcolor{cDoor}{door} that GEM labels as wall.
}
\label{fig:qual}\label{fig:pi3}
\end{figure*}
\textbf{A1: The two losses are complementary.}
The cosine anchor and relational distillation target distinct aspects of the partial-to-global
gap (Sec.~\ref{sec:why}). Tab.~\ref{tab:ablations}\,(A1) confirms this empirically:
$\mathcal{L}_{\text{cos}}$ alone reaches $63.31$ mIoU on ScanNet Partial, already strong,
since matched-point supervision transfers local class identity well, while
$\mathcal{L}_{\mathrm{KL}}$ alone reaches $55.58$ mIoU, consistent with the prediction that
relational signatures are insufficient when the absolute class direction is unconstrained.
Combining the two lifts performance to $65.46$ mIoU ($+2.15$ over cosine-only, $+9.88$ over
KL-only), and to $67.27$ with Gram ($+3.96$), confirming that the two terms capture different aspects of the alignment.\\
\textbf{A2: Random keys outperform $k$-NN keys.}
Tab.~\ref{tab:ablations}\,(A2) shows the predicted direction: replacing random sampling with k-NN-based key selection causes a substantial drop from $65.46$ to $49.71$ mIoU on ScanNet Partial, in line with the intuition that broadly sampled keys provide a more informative target distribution than spatially clustered ones.\\
\textbf{A3: Number of keys.} Performance varies by under $1.5$ mIoU across values of $M$; we use $M=4096$, the best in this sweep (details in the appendix).
\\
\textbf{Robustness to camera pitch.} Binning evaluation frames by camera pitch (Fig.~\ref{fig:pitch}), \mymethod{}-KL keeps $61$--$66$ mIoU on the steep-down, down, and level bins that make up $98.9\%$ of the data and outperforms GEM by $5$--$9$ mIoU in each. Both degrade on the rare upward views.
\begin{table}[t]
\centering
\caption{\textbf{Ablations} on ScanNet-20 partial views (Sonata, mIoU). All models use $M=4096$ random keys unless noted.}
\label{tab:ablations}
\small
\setlength{\tabcolsep}{6pt}
\begin{tabular}{llc}
\toprule
Setup & & mIoU$\uparrow$ \\
\midrule
\multicolumn{3}{l}{\textit{A1 -- Loss decomposition}} \\
$\mathcal{L}_\text{cos}$ only                        & local         & 63.31 \\
$\mathcal{L}_\text{KL}$ only                         & relational    & 55.58 \\
\ours $\mathcal{L}_\text{cos}+\mathcal{L}_\text{KL}$   & \mymethod{}-KL   & \sbest{65.46} \\
\ours $\mathcal{L}_\text{cos}+\mathcal{L}_\text{Gram}$ & \mymethod{}-Gram & \textbf{67.27} \\
\midrule
\multicolumn{3}{l}{\textit{A2 -- Key sampling (KL)}} \\
$k$-NN keys                                          & nearest in space & 49.71 \\
\ours Random keys                                    & default          & \textbf{65.46} \\
\bottomrule
\end{tabular}
\end{table}

\subsection{Application: in-the-wild segmentation}
Fig.~\ref{fig:pi3} applies \mymethod{} unchanged to point maps predicted from single ScanNet images by a feed-forward reconstruction model~\citep{vggt} (top) and to stereo point clouds from a mobile robot in unseen buildings (bottom). In both settings there is no mesh: normals are estimated from the single view with DSINE~\citep{bae2024dsine}, and no ground truth is available. \mymethod{} still produces coherent segmentations, correctly recovering the \textit{door}, while the baselines label it as wall. Acquisition details for the robot data are in Appendix~\ref{sec:robot}.

\subsection{Limitations}

\mymethod{} needs paired partial and global observations for training, which posed RGB-D provides, and transfers to new datasets without them. It inherits the frozen probe's taxonomy and accuracy, although the adapter itself is label-independent. With a calibrated IMU, gravity-aligning the frozen model's input is a strong training-free baseline for the coordinate frame (66.49 mIoU), but it cannot recover the scene context a partial view lacks. Given the same gravity, CARGO adds to it (68.27). The in-the-wild results are qualitative.

\section{Conclusion}
\label{sec:conclusion}

We presented \mymethod{}, an adaptation  framework that aligns partial camera-coordinate 3D features with a frozen global semantic space using only paired geometry. Our analysis isolates coordinate-frame mismatch as the dominant source of degradation in frozen 3D transfer, and a matched-point cosine anchor with a point-specific relational objective largely repairs it without modifying the backbone. With 1.3M trainable parameters, \mymethod{} outperforms label-supervised PEFT on ScanNet-20 with both encoders and on
all three benchmarks with Concerto, and fully fine-tuned Sonata in cross-dataset transfer.
\section{Acknowledgements}
This work was supported by the Pioneer Centre for Artificial Intelligence, funded by the Danish National Research Foundation. The authors gratefully acknowledge the computational resources provided by the P1 Compute Cluster~\cite{DTU_DCC_resource} and the Gefion AI Supercomputer, the latter supported by the Danish Centre for AI Innovation (DCAI) and the Novo Nordisk Foundation.

{
    \small
    \bibliographystyle{ieeenat_fullname}
    \bibliography{main}
}

\clearpage
\newpage
\phantomsection
\appendix

\section*{Author statement on the use of LLMs}

In this work, we used generative AI tools for writing refinement and coding assistance. They were not used for research ideation, experimental design, analysis, or scientific content. All AI-generated code was reviewed and verified by the authors before use. All research contributions are solely the responsibility of the authors.

\section{Additional Results}
\label{app:additional_res}

\subsection{Robustness across camera pose}
\label{app:pose}

Every frame in our main experiments is back-projected using its own captured
camera-to-world pose, so the $19$K training and $18$K evaluation frames already
span ScanNet's natural distribution of handheld pitch, roll, and height. (The
single-transform-to-first-frame setting appears only as the Camera|Global
diagnostic in Tab.~\ref{tab:gap} of the main paper and is not used in any main experiment.)

To characterize behaviour across this distribution, we bin evaluation frames by
camera pitch, computed from each frame's known rotation, and report
\mymethod{} against supervised GEM~\citep{gem} per bin
(Fig.~\ref{fig:pitch} of the main paper).

\mymethod{} maintains $61$--$66$ mIoU across the steep-down, downward, and level
bins, which together constitute $98.9\%$ of the evaluation set, and outperforms
supervised GEM by $5$--$9$ mIoU in each. It also improves substantially over GEM
on upward views between $10^\circ$ and $30^\circ$. Both methods degrade on the
steep-up bin, which contains only $30$ frames; the shared degradation suggests
this reflects ScanNet's limited coverage of such viewpoints rather than a failure
specific to our adaptation. Targeted augmentation for underrepresented
orientations is a natural direction for future work.

\subsection{In-the-wild segmentation}
\label{app:wild}
\textbf{Reconstructed input.}
A practical use case for our local semantic segmentation framework is compatibility with feed-forward 3D reconstruction models such as~\citep{vggt,pi32024}, which infer 3D scene attributes, including camera parameters, point maps, depth maps, and tracks, from a single RGB image.
Here, we use only the inferred point maps from ScanNet images as input.
As shown in main paper Fig.~\ref{fig:pi3} (right, top), \mymethod{} correctly identifies the \textit{door} from this limited context, without access to ground-truth depth.
By contrast, the \linearprobing{} and GEM baselines rely mainly on local planar geometry and misclassify the region as \textit{wall}.

To measure this beyond the individual examples, we build point clouds from single RGB frames of the ScanNet validation set, using either VGGT~\cite{vggt} or Depth Pro 2~\cite{DepthPro2} to estimate depth.

VGGT receives the frame resized to $504{\times}378$ (4:3, divisible by its patch size of 14), and its depth is bilinearly upsampled back to $640{\times}480$. Its output stays in VGGT's normalized units and is not rescaled to metric. Depth Pro 2 runs at the native $640{\times}480$ resolution and predicts metric depth. 

Surface normals are estimated from the same RGB frame with DSINE~\cite{bae2024dsine}, using the same intrinsics, and are oriented towards the camera. We keep points with depth in $(0.1, 8.0)$ and finite normals, then voxel-subsample at 1\,cm.

Depth, normals, and labels share one pixel grid, so each point gets the ScanNet-20 label of the pixel it was unprojected from. These per-pixel labels are the 2D annotations provided with ScanNet, which the dataset derives by projecting its 3D mesh annotations into each frame using the recorded camera poses. Pixels without a valid ScanNet-20 class are kept as ignored points.
\begin{table}[!t]
\centering
\small
\caption{\textbf{Single-image reconstructions.} Semantic segmentation on ScanNet-partial val frames, with point clouds reconstructed from a single RGB image in place of sensor depth, either as relative depth predicted by VGGT~\citep{vggt} or as metric depth predicted by Depth Pro 2~\citep{DepthPro2} and unprojected with the ScanNet intrinsics. All models are trained on ScanNet-20 Partial with sensor depth and evaluated without fine-tuning on reconstructed clouds. Best per backbone and depth source in bold.}
\label{tab:single_image}
\resizebox{\linewidth}{!}{%
\begin{tabular}{ll rrr rrr}
\toprule
& & \multicolumn{3}{c}{\textbf{ScanNet-20 VGGT}} & \multicolumn{3}{c}{\textbf{ScanNet-20 Depth Pro 2}} \\
\cmidrule(lr){3-5}\cmidrule(lr){6-8}
\textbf{Model} & \textbf{Method} & \textbf{mIoU} & \textbf{mAcc} & \textbf{allAcc} & \textbf{mIoU} & \textbf{mAcc} & \textbf{allAcc} \\
\midrule
\multirow{3}{*}{Sonata}
  & GEM              & 20.07 & 30.16 & \textbf{53.50} & 19.73 & 27.61 & 59.10 \\
  & \mymethod{}-KL   & \textbf{22.71} & \textbf{38.40} & 52.03 & \textbf{38.54} & \textbf{52.22} & \textbf{68.58} \\
  & \mymethod{}-Gram & 21.39 & 36.98 & 48.85 & 36.21 & 51.26 & 67.17 \\
\midrule
\multirow{3}{*}{Concerto}
  & GEM              & 26.77 & 40.05 & 57.99 & 33.75 & 44.97 & 67.66 \\
  & \mymethod{}-KL   & \textbf{31.58} & \textbf{52.42} & 59.11 & 49.75 & \textbf{68.92} & 76.56 \\
  & \mymethod{}-Gram & 31.40 & 46.67 & \textbf{60.23} & \textbf{50.72} & 63.11 & \textbf{76.77} \\
\bottomrule
\end{tabular}%
}
\end{table}

\begin{table}[b]
\centering
\small
\setlength{\tabcolsep}{3pt}
\caption{\textbf{Depth accuracy of reconstructed inputs.}
We evaluate predicted depth against ScanNet validation
sensor depth. VGGT predictions are treated as metric depth
without ground-truth scale alignment.}
\label{tab:depth_accuracy}
\resizebox{\columnwidth}{!}{%
\begin{tabular}{lcccccc}
\toprule
Method & AbsRel $\downarrow$
& RMSE (m) $\downarrow$
& MAE (m) $\downarrow$
& $\delta_1 \uparrow$
& $\delta_2 \uparrow$
& $\delta_3 \uparrow$ \\
\midrule
VGGT        & 0.449 & 1.12 & 1.06 & 0.108 & 0.284 & 0.518 \\
Depth Pro 2 & 0.114 & 0.32 & 0.23 & 0.917 & 0.982 & 0.992 \\
\bottomrule
\end{tabular}%
}
\end{table}
All adapters used in these experiments were trained on
ScanNet-20 Partial with the original sensor depth and are
evaluated on point clouds reconstructed from single RGB
images of the ScanNet validation set, without fine-tuning.
To characterize the quality of the reconstructed inputs,
we report depth estimation accuracy against ScanNet sensor
depth in Tab.~\ref{tab:depth_accuracy}.
Despite the substantial differences in depth accuracy,
CARGO consistently improves over GEM in mIoU across both
reconstruction sources and encoders
(Tab.~\ref{tab:depth_accuracy}).

\textbf{Stereo input on unseen scenes.}
We also evaluate stereo imagery from a mobile Husky platform in unseen environments.
Depth from stereo pairs is projected into the robot coordinate frame using known camera intrinsics. There is no ground truth but
\mymethod{} assigns plausible semantic labels to these out-of-distribution partial point clouds, producing coherent segmentation masks (main paper, Fig.~\ref{fig:pi3}, bottom).
In both settings there is no mesh, so surface normals are estimated from the single view with DSINE~\citep{bae2024dsine}.

Qualitative results were consistent across the 183 recorded frames.
Acquisition details are in Sec.~\ref{sec:robot}.

\subsection{Per-class analysis}
\label{app:perclass}

\textbf{1. Per class semantic segmentation on ScanNet-Partial (val).}
Tab.~\ref{tab:perclass_scannet_concerto} reports per-class IoU on ScanNet-Partial with the Concerto encoder. \mymethod{}-KL improves over GEM on 15 of 20 classes, with the largest gains on curtain ($+9.39$), door ($+9.22$), otherfurniture ($+7.89$), shower curtain ($+7.58$) and window ($+7.22$). GEM remains better on the bathroom classes toilet ($-12.45$), sink ($-9.51$) and bathtub ($-3.57$), and slightly on counter and sofa. \mymethod{}-Gram largely closes this gap: it improves over GEM on 17 of 20 classes and trails by at most $1.32$ IoU (toilet).

\begin{table*}[b]
\centering
\caption{\textbf{Per class IoU on \scannetegoval{}} with the Concerto encoder, for \mymethod{} trained on ScanNet-Partial. We report IoU for 20 semantic classes of ScanNet.}
\label{tab:perclass_scannet_concerto}
\resizebox{\textwidth}{!}{
\begin{tabular}{lcccccccccccccccccccc}
\toprule
& \rotatebox{55}{\textbf{wall}} & \rotatebox{55}{\textbf{floor}} & \rotatebox{55}{\textbf{cabinet}} & \rotatebox{55}{\textbf{bed}} & \rotatebox{55}{\textbf{chair}} & \rotatebox{55}{\textbf{sofa}} & \rotatebox{55}{\textbf{table}} & \rotatebox{55}{\textbf{door}} & \rotatebox{55}{\textbf{window}} & \rotatebox{55}{\textbf{bookshelf}} & \rotatebox{55}{\textbf{picture}} & \rotatebox{55}{\textbf{counter}} & \rotatebox{55}{\textbf{desk}} & \rotatebox{55}{\textbf{curtain}} & \rotatebox{55}{\textbf{refrigerator}} & \rotatebox{55}{\textbf{shower curtain}} & \rotatebox{55}{\textbf{toilet}} & \rotatebox{55}{\textbf{sink}} & \rotatebox{55}{\textbf{bathtub}} & \rotatebox{55}{\textbf{otherfurniture}} \\

\textbf{Model} &  &  &  &  &  &  &  &  &  &  &  &  &  &  &  &  &  &  &  &  \\
\midrule
Concerto (+linear) & 63.54 & 82.75 & 28.95 & 49.62 & 54.30 & 44.34 & 51.47 & 18.09 & 31.81 & 44.22 & 27.68 & 28.40 & 34.81 & 44.84 & 20.58 & 39.88 & 45.76 & 32.29 & 32.51 & 14.85 \\
\midrule
GEM & 83.37 & 95.47 & 62.48 & 80.93 & 85.10 & 77.95 & 76.19 & 61.81 & 58.88 & 71.47 & 38.68 & 57.55 & 65.14 & 69.04 & 61.31 & 63.08 & \textbf{90.38} & \textbf{65.96} & 78.99 & 54.22 \\
\mymethod{} (KL) & 84.37 & \textbf{95.50} & 66.36 & 83.81 & 86.04 & 76.53 & 80.44 & 71.03 & \textbf{66.10} & 74.68 & 41.22 & 55.38 & 67.26 & 78.43 & 63.77 & 70.66 & 77.93 & 56.45 & 75.42 & 62.11 \\
\mymethod{} (Gram) & \textbf{85.78} & 95.17 & \textbf{68.24} & \textbf{84.58} & \textbf{86.09} & \textbf{79.16} & \textbf{81.22} & \textbf{71.94} & 62.91 & \textbf{74.86} & \textbf{43.24} & \textbf{64.76} & \textbf{67.96} & \textbf{78.70} & \textbf{68.62} & \textbf{70.80} & 89.06 & 64.79 & \textbf{82.30} & \textbf{62.67} \\
$\Delta$ \mymethod{} (KL) vs GEM & \textcolor{green}{+1.00} & \textcolor{green}{+0.03} & \textcolor{green}{+3.88} & \textcolor{green}{+2.88} & \textcolor{green}{+0.94} & \textcolor{red}{-1.42} & \textcolor{green}{+4.25} & \textcolor{green}{+9.22} & \textcolor{green}{+7.22} & \textcolor{green}{+3.21} & \textcolor{green}{+2.54} & \textcolor{red}{-2.17} & \textcolor{green}{+2.12} & \textcolor{green}{+9.39} & \textcolor{green}{+2.46} & \textcolor{green}{+7.58} & \textcolor{red}{-12.45} & \textcolor{red}{-9.51} & \textcolor{red}{-3.57} & \textcolor{green}{+7.89} \\
$\Delta$ \mymethod{} (Gram) vs GEM & \textcolor{green}{+2.41} & \textcolor{red}{-0.30} & \textcolor{green}{+5.76} & \textcolor{green}{+3.65} & \textcolor{green}{+0.99} & \textcolor{green}{+1.21} & \textcolor{green}{+5.03} & \textcolor{green}{+10.13} & \textcolor{green}{+4.03} & \textcolor{green}{+3.39} & \textcolor{green}{+4.56} & \textcolor{green}{+7.21} & \textcolor{green}{+2.82} & \textcolor{green}{+9.66} & \textcolor{green}{+7.31} & \textcolor{green}{+7.72} & \textcolor{red}{-1.32} & \textcolor{red}{-1.17} & \textcolor{green}{+3.31} & \textcolor{green}{+8.45} \\
\bottomrule
\end{tabular}
}
\end{table*}

\paragraph{2. Per class cross evaluation on ScanNet++-Partial (val).} Tab.~\ref{tab:cross_scannet_scannetpp} reports per-class IoU for models trained on ScanNet-Partial and evaluated on ScanNet++-Partial. Desk, shower curtain and bathtub have no ScanNet++ points and are excluded from the mIoU (Sec.~\ref{app:mapping}). \mymethod{} improves over GEM on most classes, with large gains on bed, door, refrigerator, bookshelf, curtain, sink, and other furniture. GEM remains better on toilet, picture, and window. Label-free partial-to-global alignment thus also improves per-class transfer under dataset shift, with gains of up to $+28.11$ IoU.
\paragraph{3. Additional qualitative results} can be seen in Fig. \ref{fig:supp_qualitative}. 

\begin{table*}[ht]
\centering
\caption{\textbf{Per-class cross-evaluation (IoU) on ScanNet++-Partial (val)} for
\mymethod{} trained on ScanNet-Partial. We report IoU for the 20 ScanNet classes;
desk, shower curtain and bathtub have no ScanNet++ points (--) and are excluded, so
mIoU (last column) is averaged over the remaining 17 (Sec.~\ref{app:mapping}), matching
the cross-domain results of the main paper.}
\label{tab:cross_scannet_scannetpp}
\resizebox{\textwidth}{!}{
\begin{tabular}{lccccccccccccccccccccc}
\toprule
& \rotatebox{55}{\textbf{wall}} & \rotatebox{55}{\textbf{floor}} & \rotatebox{55}{\textbf{cabinet}} & \rotatebox{55}{\textbf{bed}} & \rotatebox{55}{\textbf{chair}} & \rotatebox{55}{\textbf{sofa}} & \rotatebox{55}{\textbf{table}} & \rotatebox{55}{\textbf{door}} & \rotatebox{55}{\textbf{window}} & \rotatebox{55}{\textbf{bookshelf}} & \rotatebox{55}{\textbf{picture}} & \rotatebox{55}{\textbf{counter}} & \rotatebox{55}{\textbf{desk}} & \rotatebox{55}{\textbf{curtain}} & \rotatebox{55}{\textbf{refrigerator}} & \rotatebox{55}{\textbf{shower curtain}} & \rotatebox{55}{\textbf{toilet}} & \rotatebox{55}{\textbf{sink}} & \rotatebox{55}{\textbf{bathtub}} & \rotatebox{55}{\textbf{otherfurniture}} & \rotatebox{55}{\textbf{mIoU}} \\

\textbf{Model} &  &  &  &  &  &  &  &  &  &  &  &  &  &  &  &  &  &  &  &  &  \\
\midrule
Sonata (full + linear) & 82.84 & 93.72 & 48.28 & 59.02 & \textbf{85.62} & 69.63 & 59.73 & 45.26 & 41.06 & 33.85 & 29.68 & \textbf{53.54} & -- & 66.97 & 35.09 & -- & \textbf{78.61} & 46.20 & -- & 32.82 & 56.58 \\
Sonata (late blocks + linear) & 75.57 & 87.73 & 31.41 & 34.75 & 70.10 & 54.37 & 48.25 & 28.07 & 28.13 & 25.90 & 14.00 & 29.95 & -- & 27.31 & 22.98 & -- & 54.22 & 34.95 & -- & 15.24 & 40.17 \\
Sonata (+linear) & 68.12 & 78.55 & 19.28 & 16.33 & 46.71 & 35.30 & 43.07 & 7.75 & 22.44 & 12.23 & 23.55 & 21.32 & -- & 33.92 & 9.34 & -- & 40.71 & 30.98 & -- & 18.54 & 31.07 \\
\midrule
GEM & 82.00 & 91.44 & 51.69 & 39.25 & 73.95 & 64.86 & 58.21 & 40.98 & \textbf{48.08} & 43.46 & \textbf{36.25} & 44.84 & -- & 63.86 & 26.84 & -- & 74.34 & 54.59 & -- & 27.29 & 54.23 \\
\mymethod{} & \textbf{88.32} & \textbf{96.14} & \textbf{58.84} & \textbf{67.36} & 83.48 & \textbf{72.71} & \textbf{60.08} & \textbf{67.61} & 46.18 & \textbf{58.48} & 33.12 & 47.41 & -- & \textbf{77.75} & \textbf{43.75} & -- & 69.34 & \textbf{65.62} & -- & \textbf{42.33} & \textbf{63.44} \\
$\Delta$ \mymethod{} vs GEM & \textcolor{green}{+6.32} & \textcolor{green}{+4.70} & \textcolor{green}{+7.15} & \textcolor{green}{+28.11} & \textcolor{green}{+9.53} & \textcolor{green}{+7.85} & \textcolor{green}{+1.87} & \textcolor{green}{+26.63} & \textcolor{red}{-1.90} & \textcolor{green}{+15.02} & \textcolor{red}{-3.13} & \textcolor{green}{+2.57} & -- & \textcolor{green}{+13.89} & \textcolor{green}{+16.91} & -- & \textcolor{red}{-5.00} & \textcolor{green}{+11.03} & -- & \textcolor{green}{+15.04} & \textcolor{green}{+9.21} \\
\bottomrule
\end{tabular}
}
\end{table*}

\subsection{Relational loss parameterization}
\label{app:relational}

The relational objective constrains each partial feature's similarity pattern
to a set of global reference keys (main paper, Sec.~\ref{sec:loss}).
Tab.~\ref{tab:relational} (and Tab.~\ref{tab:partial_view_all} of the main paper) compares two parameterizations of this constraint:
KL divergence between softmax distributions (\mymethod{}, the default), and direct regression of the
sampled similarities (\mymethod{}-Gram).

Both improve substantially over the cosine anchor alone, and on ScanNet-20 the choice between
them is a trade-off rather than a ranking. \mymethod{}-Gram gives higher mIoU
($+1.81$ on Sonata, $+2.47$ on Concerto) and higher allAcc, while KL gives
higher mAcc ($+2.75$ and $+2.51$ respectively). Since mAcc weights all classes
equally regardless of frequency, the KL formulation is preferable when
per-class balance matters, and \mymethod{}-Gram when overall segmentation
quality is the target. We use KL as the default variant and for all ablations, and report \mymethod{}-Gram alongside it in the main paper. On the finer taxonomies the ranking is not fixed: KL is stronger on ScanNet++ with both encoders, whereas Gram is stronger on ScanNet200 (Tab.~\ref{tab:partial_view_all} of the main paper). That the ScanNet-20 pattern is consistent across two
architecturally distinct encoders supports our claim that the benefit derives
from point-specific relational supervision rather than from a specific
divergence.

\paragraph{Number of relational keys.}
We additionally investigate the effect of the
number of randomly sampled global reference keys
$M$ (Tab.~\ref{tab:ablation_keys}).
Performance on ScanNet Partial varies by less than
1.5 mIoU across $M \in \{2048,4096,8192\}$,
with $M=4096$ achieving the best performance
(65.46 mIoU). The same configuration also
achieves the highest zero-shot cross-domain
performance on ScanNet++ Partial (53.93 mIoU).

\begin{table}[!t]
\centering
\small
\caption{\textbf{Relational loss parameterization} on ScanNet-20 Partial (val).
\mymethod{}-KL is the default variant, denoted \mymethod{} in the main paper;
\mymethod{}-Gram regresses the sampled similarities directly (main paper, Sec.~\ref{sec:loss}).}
\label{tab:relational}
\begin{tabular}{llrrr}
\toprule
\textbf{Model} & \textbf{Variant} & \textbf{mIoU} & \textbf{mAcc} & \textbf{allAcc} \\
\midrule
\multirow{2}{*}{Sonata}
  & \mymethod{}-KL   & 65.46 & \textbf{80.50} & 85.93 \\
  & \mymethod{}-Gram & \textbf{67.27} & 77.75 & \textbf{87.62} \\
\midrule
\multirow{2}{*}{Concerto}
  & \mymethod{}-KL   & 71.67 & \textbf{87.77} & 89.96 \\
  & \mymethod{}-Gram & \textbf{74.14} & 85.26 & \textbf{90.66} \\
\bottomrule
\end{tabular}
\end{table}

\subsection{ScanNet++ to ScanNet-20 class mapping}
\label{app:mapping}
For cross-dataset evaluation we use ScanNet++'s benchmark mapping and the ScanNet v2 category table: 26 ScanNet++ classes map to 17 ScanNet-20 classes, 11 one-to-one plus the merges wall + shower wall, cabinet + kitchen cabinet, chair + office chair, door + doorframe, picture + poster + painting, and otherfurniture = heater, trash can, crate, rack. All other ScanNet++ classes are ignored. Desk, shower curtain and bathtub have no ScanNet++ points and are excluded from the mIoU, which is therefore averaged over 17 classes.

\subsection{Sensitivity to gravity error}
\label{app:gravity}

An alternative to learned adaptation is to rotate each partial view into a
gravity-aligned frame and apply the frozen, globally trained model. We
evaluate this alternative with the frozen Sonata model and the global
probe. Each ScanNet Partial frame is gravity-aligned using its ground-truth
camera tilt, perturbed by zero-mean Gaussian noise with standard deviation
$\sigma$ on each tilt axis (Tab.~\ref{tab:gravity_noise}).

\paragraph{Oracle gravity.}
With exact gravity, the frozen model reaches 66.49 mIoU. This remains
below the 67.45 mIoU it obtains on global-frame input (main paper, Tab.~\ref{tab:gap}):
gravity alignment corrects the orientation of the view, but not its
camera-centred origin and height. The oracle is also favourable to the
frozen model. ScanNet scenes are stored in a gravity-aligned frame with
$z$ pointing up, a convention that the pretrained encoder and the global
probe implicitly rely on. Aligning the input to gravity restores this
dataset-specific convention at test time, rather than addressing the
shift to camera-frame input in general.

\paragraph{Noisy gravity.}
Accuracy is stable for $\sigma \leq 2^\circ$ but degrades beyond $5^\circ$,
reaching 61.52 mIoU at $\sigma = 10^\circ$. Errors of only a few degrees
require a precise gravity estimate. An IMU provides one reliably only when
the platform is close to static: on handheld and mobile platforms,
accelerations corrupt the estimate exactly when the camera is moving. For
single-image reconstructions (Appendix~\ref{app:wild}), no sensor-based estimate is
available at all. Image-based gravity estimation~[4] has a median tilt
error of $5.8^\circ$ on these frames, with 20.2\% of frames exceeding
$10^\circ$, and yields 41.64 mIoU (main paper, Tab.~\ref{tab:gap}).

\paragraph{Comparison with CARGO.}
Unlike gravity-aligned baselines, CARGO operates directly on camera-frame inputs without requiring gravity estimation or alignment at inference time. CARGO-KL achieves 65.46 mIoU, approaching the ground-truth gravity-aligned frozen model (66.49), while CARGO-Gram reaches 67.27 mIoU, exceeding this oracle baseline. Both substantially outperform the frozen model using estimated gravity (41.64 mIoU).

\begin{table}[t]
\centering
\small
\setlength{\tabcolsep}{7pt}
\caption{
\textbf{Robustness to gravity alignment noise}
on ScanNet Partial.
Frozen Sonata is evaluated over three random yaw
rotations with gravity tilt perturbations of
standard deviation $\sigma$. (GT only random yaw.)
Results are mean $\pm$ standard dev.
CARGO requires no gravity estimation or
alignment at inference time.
}
\label{tab:gravity_noise}
\begin{tabular}{lc}
\toprule
Method / Gravity alignment & mIoU (\%) \\
\midrule
\multicolumn{2}{l}{\textit{Frozen Sonata}} \\
GT gravity ($\sigma=0^\circ$)
    & $66.49 \pm 0.04$ \\
$\sigma=1^\circ$ & $66.57 \pm 0.01$ \\
$\sigma=2^\circ$ & $66.57 \pm 0.03$ \\
$\sigma=5^\circ$ & $66.21 \pm 0.02$ \\
$\sigma=7.5^\circ$ & $64.83 \pm 0.02$ \\
$\sigma=10^\circ$ & $61.52 \pm 0.14$ \\
Estimated gravity~\citep{persp-fields}
    & $41.64$ \\
\midrule
\multicolumn{2}{l}{
\textit{Ours (no gravity estimation)}} \\
CARGO-KL & $65.46$ \\
CARGO-Gram & $\mathbf{67.27}$ \\
\bottomrule
\end{tabular}
\end{table}

\definecolor{clsWall}    {RGB}{174,199,232}
\definecolor{clsFloor}   {RGB}{152,223,138}
\definecolor{clsCabinet} {RGB}{ 31,119,180}
\definecolor{clsBed}     {RGB}{255,187,120}
\definecolor{clsChair}   {RGB}{188,189, 34}
\definecolor{clsSofa}    {RGB}{140, 86, 75}
\definecolor{clsTable}   {RGB}{255,152,150}
\definecolor{clsDoor}    {RGB}{214, 39, 40}
\definecolor{clsWindow}  {RGB}{197,176,213}
\definecolor{clsShelf}   {RGB}{148,103,189}
\definecolor{clsPicture} {RGB}{196,156,148}
\definecolor{clsCounter} {RGB}{ 23,190,207}
\definecolor{clsDesk}    {RGB}{247,182,210}
\definecolor{clsCurtain} {RGB}{219,219,141}
\definecolor{clsShelves} {RGB}{ 66,188,102}
\definecolor{clsDresser} {RGB}{140, 57,197}
\definecolor{clsClothes} {RGB}{ 92,193, 61}
\definecolor{clsPillow}  {RGB}{202,185, 52}
\definecolor{clsMirror}  {RGB}{ 51,176,203}
\definecolor{clsFloormat}{RGB}{200, 54,131}
\definecolor{clsIgnore}  {RGB}{  0,  0,  0}

\newcommand{\clsdot}[2]{\mbox{\textcolor{#1}{\large$\bullet$}\,{\small #2}}%
                        \hspace{0.9em plus 0.4em minus 0.25em}}

\begin{figure*}[t]
  \centering
  \noindent%
  \makebox[0.2\linewidth][c]{Input}%
  \makebox[0.2\linewidth][c]{\textit{Sonata (linear)}}%
  \makebox[0.2\linewidth][c]{GEM}%
  \makebox[0.2\linewidth][c]{\textbf{CARGO}}%
  \makebox[0.2\linewidth][c]{GT}\\[2pt]

  \includegraphics[width=\linewidth]{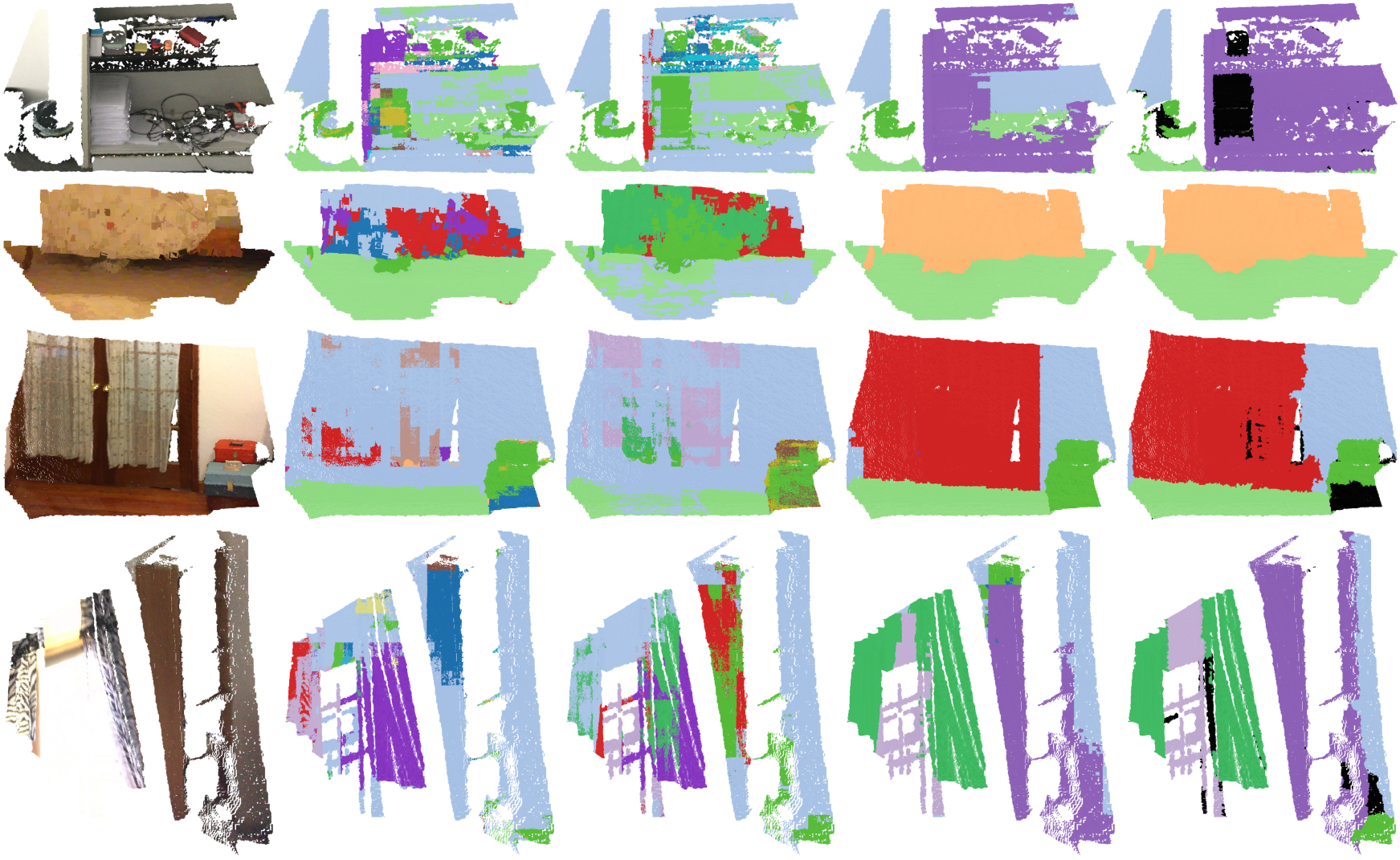}\\[5pt]

  \parbox{\linewidth}{\centering
    \clsdot{clsWall}{wall}\clsdot{clsFloor}{floor}\clsdot{clsBed}{bed}%
    \clsdot{clsDoor}{door}\clsdot{clsWindow}{window}\clsdot{clsShelf}{bookshelf}%
    \clsdot{clsCabinet}{cabinet}\clsdot{clsCounter}{counter}%
    \clsdot{clsPicture}{picture}\clsdot{clsDesk}{desk}\clsdot{clsTable}{table}%
    \clsdot{clsSofa}{sofa}\clsdot{clsChair}{chair}\clsdot{clsCurtain}{curtain}%
    \clsdot{clsShelves}{shelves$^{}$}\clsdot{clsDresser}{dresser$^{}$}%
    \clsdot{clsClothes}{clothes$^{}$}\clsdot{clsPillow}{pillow$^{}$}%
    \clsdot{clsMirror}{mirror$^{}$}\clsdot{clsFloormat}{floor mat$^{}$}%
    \clsdot{clsIgnore}{unlabeled}}

  \caption{\textbf{Additional qualitative results on ScanNet Partial.}
  Each row is a single camera-frame partial view.
  \textit{Row 1:} both baselines label the shelving unit predominantly
  \emph{wall}, with \textit{Sonata} adding \emph{dresser} and \emph{clothes}.
  CARGO recovers it as a single \emph{bookshelf} %
  \textit{Row 2:} the baselines fragment the bed into \emph{floor},
  \emph{door}, \emph{wall} and \emph{shelves}, while CARGO reproduces the
  ground-truth almost exactly. %
  \textit{Row 3:} neither baseline predicts \emph{door} anywhere, labelling the
  surface \emph{wall}. CARGO recovers it. 
  \textit{Row 4:} CARGO tracks the ground truth closely on \emph{shelves},
  whereas the
  baselines over-predict \emph{wall}.
  Unannotated points are shown in black.}
  \label{fig:supp_qualitative}
\end{figure*}

\subsection{Comparison with D-DITR}
\label{app:ditr}
We apply the public D-DITR~\citep{knaebel2026ditr} checkpoints, trained on ScanNet-20 and ScanNet-200 labels, directly to partial-view inputs (Tab.~\ref{eval_ditr}). We evaluate both ScanNet Partial and its 2.5D counterpart.
D-DITR was not designed for camera-coordinate partial-view inference, so the gap primarily reflects this input shift rather than a like-for-like algorithmic comparison; it indicates that a model trained in global coordinates, even with 2D-to-3D feature distillation, does not transfer to camera-coordinate partial views without adaptation.

\begin{table}[!htbp]
\centering
\scriptsize
\setlength{\tabcolsep}{5pt}
\renewcommand{\arraystretch}{1.05}
\caption{\textbf{Evaluation with D-DITR.} We evaluate both ScanNet Partial and its RGB-D reconstruction (ScanNet Partial-2.5D), using D-DITR\,\citep{knaebel2026ditr}. We report results using two publicly available D-DITR variants trained on ScanNet-20 and ScanNet-200 labels respectively.}
\label{eval_ditr}
\begin{tabular}{l rr rr}
\toprule
\multirow{2}{*}{\textbf{Method}} &
\multicolumn{2}{c}{\textbf{ScanNet-20  } (val)} &
\multicolumn{2}{c}{\textbf{ScanNet-200} (val) } \\
\cmidrule(lr){2-3} \cmidrule(lr){4-5}
& \textbf{Partial} mIoU & \textbf{2.5D} mIoU & \textbf{Partial} mIoU & \textbf{2.5D} mIoU \\
\midrule
D-DITR \citep{knaebel2026ditr}
& 1.70 & 1.66 & 0.29 & 0.24 \\
\mymethod{}
& \textcolor{green!50!black}{\tiny(+63.76)}\,\textbf{65.46}
& \textcolor{green!50!black}{\tiny(+62.22)}\,\textbf{63.88}
& \textcolor{green!50!black}{\tiny(+18.96)}\,\textbf{19.25}
& \textcolor{green!50!black}{\tiny(+16.88)}\,\textbf{17.12} \\
\bottomrule
\end{tabular}
\end{table}

\subsection{Choice of encoder}
\label{app:encoder}
\begin{table}[!b]
\centering
\caption{\textbf{Globally trained} PTv3 vs.\ Sonata (Decoder) on the partial setup.}
\begin{tabular}{lccc}
\toprule
Method & mIoU & mAcc & allAcc \\
\midrule
PTv3   & 1.30 & 5.03 & 14.28 \\
Sonata (Decoder) & \textbf{2.60} & \textbf{6.20} & \textbf{24.14} \\
\bottomrule
\end{tabular}
\label{tab:ptv3}
\end{table}
In addition to our experiments with Sonata and Concerto encoders, we compare the Sonata encoder with PTv3~\citep{pointtransformerv3}, another widely used 3D backbone. Without adaptation, both globally trained models collapse on partial camera-coordinate input (Tab.~\ref{tab:ptv3}), with Sonata slightly ahead ($2.60$ vs.\ $1.30$ mIoU). Since PTv3 uses a different decoder, we also evaluate Sonata with the PTv3 decoder for a controlled comparison. We therefore use Sonata, alongside Concerto, for all main experiments.

\section{Model Details}
\label{app:model_details}

\subsection{Implementation details}
\label{sec:more_details}
Our implementation is based on the open-source Pointcept codebase and follows
the official implementations of Sonata~\citep{sonata} and GEM~\citep{gem}.

For a fair comparison, we train all baselines for 100 epochs, following the GEM
training protocol~\citep{gem}. We found that all baselines had converged by this
point. We also used the same batch size. Tab.~\ref{tab:details_bridge3d} lists the \mymethod{} training configuration.\\
\textbf{Software and hardware environment.} All experiments used a single NVIDIA H100 PCIe GPU with 80 GB memory for training and testing (Tab.~\ref{tab:env}).
\begin{table}[!htbp]
    \centering
    \caption{\textbf{Hardware and software environment} used for all experiments.}
    \small
    \resizebox{0.8\columnwidth}{!}{%
    \begin{tabular}{l l}
        \toprule
        \textbf{Component} & \textbf{Specification} \\
        \midrule
        Operating system & AlmaLinux release 9.6 \\
        CPU & AMD EPYC 9354 32-Core \\
        RAM & 768GB \\
        GPU & NVIDIA H100 PCIe 80GB \\
        CUDA version (system) & 13.0 \\
        Python version & 3.12.9 \\
        PyTorch version & 2.5.1+cu124 \\
        \bottomrule
    \end{tabular}%
    }
    \label{tab:env}
\end{table}

\section{Dataset and Preprocessing Details}
\label{app:dataset_preprocessing}

For each RGB-D frame, we obtain a depth map $D$, color image, camera intrinsics $K$, and camera-to-world pose $T_{\text{cam} \to \text{world}}$$=[\mathbf{R}\,|\,\mathbf{t}]$. For every valid pixel $p = (u,v)$ with depth $D_p$, we backproject to 3D in the camera frame,

\begin{equation}
    p^{\text{cam}} = D_p\,\mathbf{K}^{-1}\tilde{p},
\end{equation}
where $\tilde{p}=(u,v,1)^\top$ is the pixel in homogeneous coordinates.

We discard invalid depths (non-finite, zero, or beyond a maximum range) and transform the remaining points to the global coordinate system using

\begin{equation}
    p^{\text{world}}
        = \mathbf{R} \bigl(D_p\,\mathbf{K}^{-1}\tilde{p}\bigr) + \mathbf{t}.
\end{equation}

The RGB values are sampled at $(u,v)$ from the color image. This produces, for each frame, a dense partial point cloud.

To make partial point clouds consistent with the original ScanNet annotations, we project semantics and normals from the cleaned mesh onto the RGB-D points. For each scene, we load the corresponding mesh and compute vertex normals. Using the dataset's labels we assign each vertex semantic and instance IDs.

We then establish correspondences between mesh vertices and RGB-D points in world coordinates by building a KD-tree over the frame points and querying it with all valid mesh vertices. A mesh vertex and frame point form a pair if their Euclidean distance $\leq 0.01\,\text{m}$. For each RGB-D point, we keep only the nearest valid mesh vertex within this threshold. RGB-D points without any valid correspondence are discarded.

For each retained point, we copy the semantic and instance labels as well as the vertex normal from the matched mesh vertex. Normals are rotated into the camera frame as follows

\begin{equation}
n^{\text{cam}} = \mathbf{R}^\top n^{\text{world}}.
\end{equation}

\subsection{Correspondences between Partial Point clouds and the Global Mesh}
The same world-space coordinates are also used to construct the voxel-level correspondences used in our loss. For each scene, we perform a nearest-neighbor search in the global space and assign each original partial point cloud coordinate to its closest global scene original coordinate.

To ensure these correspondences remain valid after augmentation, we explicitly track how every point moves throughout the entire transform chain. At the beginning of preprocessing, each point is assigned a persistent origin identifier, and after every point-indexing operation we update the mapping that links the current points back to their original positions. During voxelization, we additionally record which original points correspond to each voxel representative, allowing us to propagate origin-level correspondences into voxel-level correspondences that remain aligned with the subsequent augmentations.

\subsection{Mobile Robot Data Acquisition}
\label{sec:robot}
To assess \mymethod{} under real-world conditions, we collect partial point clouds using a Clearpath Husky A200 UGV equipped with a Roboception RC-Visard 160c stereo camera mounted on a rigid front-facing tower. The camera provides synchronized left/right RGB images and onboard SGM-based disparity and depth maps, which we use exclusively.

Camera intrinsics are obtained through Roboception's built-in calibration tool, and the stereo system outputs rectified images, so no additional geometric calibration is required. We back-project the stereo depth into 3D using the calibrated intrinsics to form partial point clouds.

Points beyond $7$~m are discarded to remove distant background measurements.
Surface normals are estimated with~\citep{bae2024dsine}.

\subsection{Dataset Statistics}
\label{sec:appendix}

\textbf{Datasets.} We employ three widely used indoor 3D datasets: ScanNet~\citep{scannet}, ScanNet200~\citep{rozenberszki2022language} and ScanNet++~\citep{scannetpp}: (1) ScanNet is collected using a lightweight RGB-D scanning setup and comprises $1513$ indoor scenes, covering approximately $2.5\text{M}$ views, with corresponding 3D camera poses, surface reconstructions, and point-level semantic annotations for 20 classes and 200 classes (for the ScanNet200 version respectively).
For training all methods we sample every 100\textsuperscript{th} frame from the RGB-D sequences ($\sim 0.3$ FPS) of the training set, resulting in approximately $19\text{K}$ training frames. This higher sampling stride was chosen to ensure training efficiency given the large number of available frames.
For evaluation we sample every 30\textsuperscript{th} frame ($\sim 1$ FPS) from the validation set, resulting in about $18\text{K}$ evaluation frames. (2) We also use ScanNet++, which comes with
high-fidelity 3D mask annotations, including smaller objects (100 classes). It includes high-resolution 3D scans captured at sub-millimeter precision
and annotated comprehensively, covering objects of varying sizes.

For evaluation we sample every 60\textsuperscript{th} frame ($\sim 1$ FPS) from the validation set, resulting in about $7\text{K}$ evaluation frames. All methods are trained and evaluated on the same subsets for fair comparison (see Tab.~\ref{tab:datasets} for details).

\begin{table}[t]
\centering
\vspace{1mm}
\caption{\textbf{Dataset statistics.} 
Number of scenes and average points per scene.
Partial variants contain multiple partial reconstructions per scene therefore, we sample /100 for training on ScanNet and /60 for training on ScanNet++. Evaluation follows /30 in ScanNet and /60 in ScanNet++.}
\scriptsize
\setlength{\tabcolsep}{2pt}
\renewcommand{\arraystretch}{1.1}
\begin{tabular}{l|c|rrr|rr|rr}
\toprule
Dataset & Src &
\multicolumn{3}{c|}{\textbf{Scenes}} &
\multicolumn{2}{c|}{$\overline{\text{pts/scene}}$} &
\multicolumn{2}{c}{\textbf{Overall}} \\
\cmidrule(lr){3-5}\cmidrule(lr){6-7}\cmidrule(lr){8-9}
& & Train & Val & Test & Train & Val & Scenes & Points \\
\midrule
ScanNet~\citep{scannet} & real &
1.2K & 312 & 100 & 145K & 159K & 1.6K & 242M \\
ScanNet-Partial & real &
19.6K & 17.8K & N/A & 6.5K & 6.6K & 37.4K & 244M \\
\midrule
ScanNet++~\citep{scannetpp} & real &
856 & 50 & 50 & 1.50M & 1.51M & 956 & 1.43B \\
ScanNet++-Partial & real &
38.4K & 7.1K & N/A & 22.7K & 21.6K & 45.5K & 1.03B \\
\bottomrule
\end{tabular}

\label{tab:datasets}
\end{table}
For our cross-dataset experiments, we use the ScanNet training protocol, but we evaluate on ScanNet++ validation.

\begin{table}[t]
\centering
\caption{\textbf{Number of relational keys $M$.}
mIoU on ScanNet Partial (in-domain) and
ScanNet++ Partial (zero-shot cross-eval).
All models are trained on ScanNet Partial.
Percentages denote the fraction of global scene
points used as relational keys.}
\label{tab:ablation_keys}
\small
\setlength{\tabcolsep}{6pt}
\begin{tabular}{lcc}
\toprule
\textbf{Keys $M$} &
\textbf{ScanNet} &
\textbf{ScanNet++} \\
\midrule
2048 ($1.4\%$) & 64.01 & 50.80 \\
\textbf{4096 ($2.8\%$)} & \textbf{65.46} & \textbf{53.93} \\
8192 ($5.6\%$) & 64.63 & 53.19 \\
\bottomrule
\end{tabular}
\end{table}

\subsection{Broader Impacts}
Reliable partial-view 3D segmentation supports applications such as robotics, AR/VR, and assistive navigation, where embodied agents must reason from limited onboard observations. As with any indoor system, deployment in human-centred environments raises privacy and monitoring considerations that warrant care in real-world use.

\begin{table*}[b]
\caption{\mymethod{} training configuration for ScanNet Partial and ScanNet++ Partial.}
\centering
\small
\label{tab:details_bridge3d}
\begin{tabular}{lcc}
\toprule
\textbf{Configuration} & \textbf{ScanNet-20/200 Partial} & \textbf{ScanNet++ Partial} \\
\midrule

\multicolumn{3}{l}{\textbf{Optimization}} \\
Optimizer & AdamW & AdamW \\
Learning rate & $2 \times 10^{-5}$ & $2 \times 10^{-5}$ \\
LR schedule & Constant & Constant \\
Weight decay & $2 \times 10^{-2}$ & $2 \times 10^{-2}$ \\
Gradient clipping & 3.0 & 3.0 \\
Mixed precision & FP16 (AMP) & FP16 (AMP) \\
Batch size & 24 & 24 \\
Epochs & 100 & 100 \\
Trainable modules & spatial adapter only & spatial adapter only \\

\midrule
\multicolumn{3}{l}{\textbf{Backbone, frozen}} \\
Backbone & PT-v3m2, Sonata/Concerto weights & PT-v3m2, Sonata/Concerto weights \\
Input channels & 9, xyz + rgb + normal & 9, xyz + rgb + normal \\
Encoder depths & (3, 3, 3, 12, 3) & (3, 3, 3, 12, 3) \\
Encoder channels & (48, 96, 192, 384, 512) & (48, 96, 192, 384, 512) \\
Encoder heads & (3, 6, 12, 24, 32) & (3, 6, 12, 24, 32) \\
Patch size & 1024, all stages & 1024, all stages \\
Order & (z, z-trans, hilbert, hilbert-trans) & (z, z-trans, hilbert, hilbert-trans) \\
Voxel size & 0.02 m & 0.02 m \\
Drop path & 0.3 & 0.3 \\
Attention impl. & FlashAttention & FlashAttention \\

\midrule
\multicolumn{3}{l}{\textbf{Adapter}} \\
Spatial adapter hidden dim & 32 & 32 \\
Spatial projection & lin-subm3 & lin-subm3 \\
Adapter init & LoRA-style & LoRA-style \\

\midrule
\multicolumn{3}{l}{\textbf{\mymethod{} loss}} \\
Student temperature $\tau_s$ & 0.10 & 0.10 \\
Teacher temperature $\tau_t$ & 0.04 & 0.04 \\
Number of relational keys $M$ & 4096 & 16384 \\

\midrule
\multicolumn{3}{l}{\textbf{Data}} \\
Frame stride, train & 100 & 60 \\
Frame stride, eval & 30 & 60 \\

\bottomrule
\end{tabular}
\end{table*}

\end{document}